\documentclass[11pt]{article}       

\usepackage{a4}

\usepackage{graphicx}
\usepackage{mathptmx}      
\usepackage{amssymb,amsmath,amsfonts}
\usepackage{multirow}

\begin{document}

\title{Classifying pairs with trees \\for supervised biological network inference
}


\author{Marie~Schrynemackers$^1$ \and Louis~Wehenkel$^1$ \and M.~Madan~Babu$^2$ \and Pierre~Geurts$^1$\\
$^1$Department of EE and CS \& GIGA-R, University of Li\`{e}ge, Belgium\\
$^2$MRC Laboratory of Molecular Biology, Cambridge, UK}



\date{\today}

\maketitle

\begin{abstract}
Networks are ubiquitous in biology and computational approaches have 
been largely investigated for their inference. In particular, supervised machine learning methods can be used to
complete a partially known network by integrating various
measurements. Two main supervised frameworks have been proposed: the
local approach, which trains a separate model for each network node,
and the global approach, which trains a single model over pairs of
nodes. Here, we systematically investigate, theoretically and
empirically, the exploitation of tree-based ensemble methods in the
context of these two approaches for biological network inference.
We first formalize the problem of network inference as classification
of pairs, unifying in the process homogeneous and bipartite graphs and
discussing two main sampling schemes. We then present the global and
the local approaches, extending the later for the prediction of
interactions between two unseen network nodes, and discuss their
specializations to tree-based ensemble methods, highlighting their
interpretability and drawing links with clustering
techniques. Extensive computational experiments are carried out with
these methods on various biological networks that clearly highlight
that these methods are competitive with existing methods.
\end{abstract}

\section{Introduction} \label{intro}

In biology, relationships between biological entities (genes,
proteins, transcription factors, micro-RNA, diseases, etc.) are often
represented by graphs (or networks\footnote{In this paper, the terms
  network and graph will refer to the same thing.}). In theory, most
of these networks can be identified from lab experiments but in
practice, because of the difficulties in setting up these experiments
and their costs, we often have only a very partial knowledge of
them. Because more and more experimental data become available about
biological entities of interest, several researchers took an interest
in using computational approaches to predict interactions between
nodes in order to complete experimental predictions.

When formulated as a supervised learning problem, network inference
consists in learning a classifier on pairs of nodes. Mainly two
approaches have been investigated in the literature to adapt existing
classification methods for this problem \cite{vert2009}. The first one,
that we call the global approach, considers this problem as a standard
classification problem on an input feature vector obtained by
concatenating the feature vectors of each node from the pair
\cite{vert2009}. The second approach, called local \cite{bleakley2007,mordelet2008},
trains a different classifier for each node separately, aiming at
predicting its direct neighbors in the graph. These two approaches
have been mainly exploited with support vector machine (SVM)
classifiers. In particular, several kernels have been proposed for
comparing pairs of nodes in the global approach
\cite{benhur2005,vert2007} and the global and local approaches can be
related for specific choices of this kernel \cite{hue2010-2}. A
number of papers applied the global approach with tree-based ensemble
methods, mainly Random Forests \cite{breiman2001}, for the prediction of
protein-protein \cite{lin2004,chen2005,qi2006,tastan2009} and
drug-protein \cite{yu2012} interactions, combining various feature
sets. Besides the local and global methods, other approaches for
supervised graph inference includes, among others, matrix completion
methods \cite{kato2005}, methods based on output kernel regression
\cite{geurts2007,brouard2011}, Random Forests-based similarity
learning \cite{qi2005}, and methods based on network properties
\cite{cheng2012}.

In this paper, we would like to systematically investigate,
theoretically and empirically, the exploitation of tree-based ensemble
methods in the context of the local and global approaches for
supervised biological network inference. We first formalize biological
network inference as the problem of classification on pairs,
considering in the same framework homogeneous graphs, defined on one
kind of nodes, and bipartite graphs, linking nodes of two families. We
then define the general local and global approaches in the context of
this formalization, extending in the process the local approach for
the prediction of interactions between two unseen network nodes. The
paper discusses in details the specialization of these approaches to
tree-based ensemble methods. In particular, we highlight their high
potential in terms of interpretability and draw 
connections between these methods and unsupervised (bi-)clustering
methods. Experiments on several biological networks show the good
predictive performance of the resulting family of methods. Both the
local and the global approaches are competitive with however an advantage for the local approach in terms of
compactness of the inferred models.

The paper is structured as follows. Section~\ref{sec:classpair}
defines the general problem of classification on pairs and discusses
two different sampling protocols in this
context. Section~\ref{methods} presents the global and local
approaches and their particularization for tree
ensembles. Section~\ref{sec:experiments} relates experiments with
these methods on several homogeneous and bipartite biological
networks. Section~\ref{concl} concludes and discusses future work
directions. Additional results can be found in the appendix.


\section{Network inference as classification on pairs}\label{sec:classpair}

For the sake of generality, we assume that we have two finite sets of nodes, $U_r=\{n_r^1,\ldots,n_r^{N_{U_r}}\}$ and $U_c=\{n_c^1,\ldots,n_c^{N_{U_c}}\}$. An adjacency matrix $Y$ of size $N_{U_r} \times N_{U_c}$ can then define a network connecting the two sets of nodes. An entry $y_{ij}$ is equal to one if there is an edge between the nodes $n_r^i$ and $n_c^j$, and zero if not. The subscripts $r$ and $c$ stand respectively for \textit{row} and \textit{column} of the adjacency matrix $Y$. Moreover, each node (or sometimes pair of nodes) is described by a feature representation given by $x(n)$ (or $x(n_r,n_c)$ for a pair), typically lying in $\mathbb{R}^p$. $Y$ thus defines a partial \textit{bipartite} graph over the two sets $U_r$ and $U_c$. \textit{Homogeneous} graphs defined on only one family of nodes can nevertheless be obtained as special cases of this general framework by considering only one universe of nodes ($U=U_r =U_c$). \cite{schrynemackers2013}

In this context, the problem of network inference can be cast as a problem of classification on pairs:
\begin{quote}
Given a partial knowledge of the adjacency matrix $Y$ of the target
network in the form of a learning sample of triplets:
$$LS_p=\{(n_r^{i_k},n_c^{j_k},y_{i_kj_k})|k=1,\ldots,N_{LS}\},$$ and given the
feature representation of the nodes and/or pairs of nodes, find a function
$f:U_r\times U_c\rightarrow \{0,1\}$ that best approximates the unlabeled entries
of the adjacency matrix from their feature representation (on nodes or on pairs).
\end{quote}

\begin{figure}
\centering
\includegraphics[width=10cm]{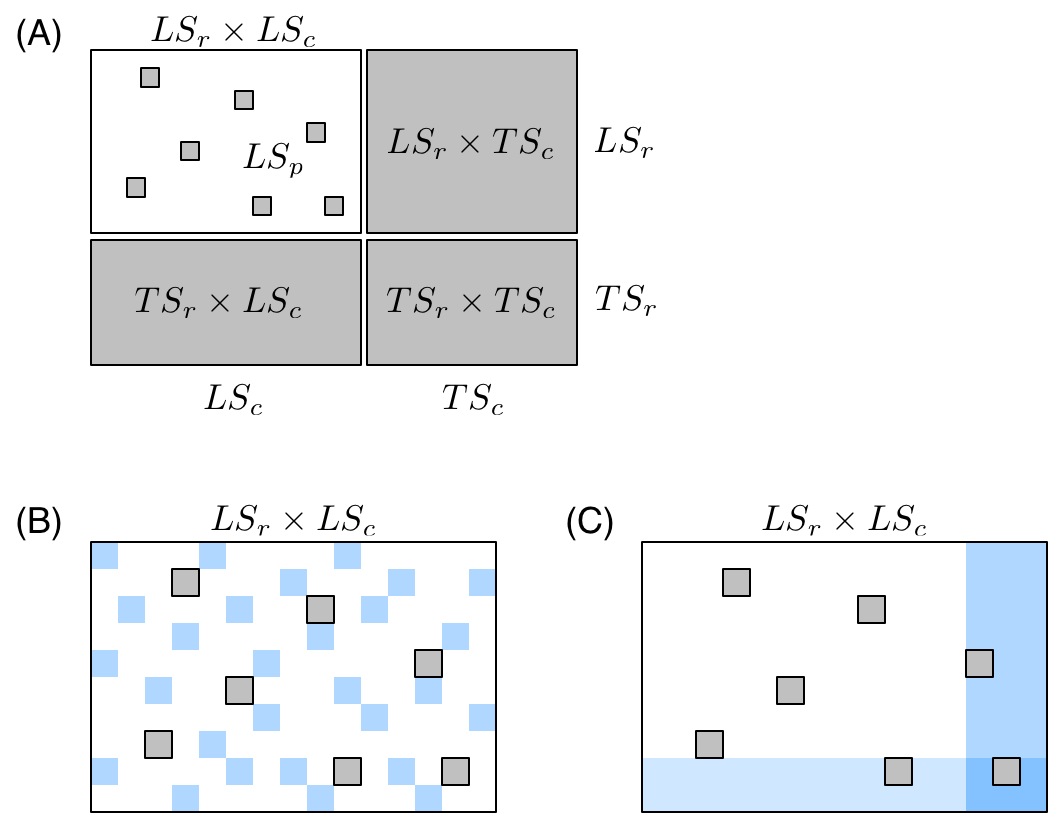}
\caption{Schematic representation of known and unknown pairs in the
    network adjacency matrix (A) and of the two kinds of CV,
    CV on pairs (B) and CV on nodes (C). In (A):
    known pairs (that can be interacting or not) are in white and unknown
    pairs, to be predicted, are in gray. Rows and columns of the adjacency
    matrix have been rearranged to highlight the four families of unknown pairs
    described in the text: $LS_r\times LS_c$, $LS_r\times TS_c$, $TS_r\times LS_c$, and
    $TS_r\times TS_c$. In (B),(C): pairs from the learning fold are in white and
    pairs from the test fold are in blue. Pairs in gray represent unknown pairs
    that do not take part to the CV.}
\label{protocols2}
\end{figure}

Given a learning set $LS_p$ of pairs labeled as interacting or not, the goal of a supervised network inference method is to get a prediction for the pairs not present in $LS_p$. All the pairs are not as easy as the others to predict: it is typically much more difficult to predict pairs with nodes for which no example of interactions are provided in the training network. One may thus partitioned the prediction into four families, depending on whether the nodes in the tested pair are represented or not in the learning set $LS_p$. Denoting by
$LS_c$ (resp. $LS_r$) the nodes from ${U}_c$ (resp. ${U}_r$) that are present
in $LS_p$ (i.e. which are involved in some pairs in $LS_p$) and by
$TS_c={U}_c\setminus LS_c$ (resp. $TS_r={U}_r\setminus LS_r$) unseen nodes from
${U}_c$ (resp. ${U}_r$), the pairs of nodes to predict (i.e., outside $LS_p$)
can be divided into the following four families:
\begin{itemize}
\item $(LS_r \times LS_c) \setminus LS_p $: predictions of (unseen) pairs
  between two nodes which are represented in the learning sample.
\item $LS_r \times TS_c$ or $TS_r \times LS_c$: predictions of pairs between
  one node represented in the learning sample and one unseen node, where the
  unseen node can be either from $U_c$ or from ${U}_r$.
\item $TS_r \times TS_c$: predictions of pairs between two unseen nodes.
\end{itemize}
These families of pairs are represented in the adjacency matrix in Figure~\ref{protocols2}(A). Thereafter, we denote the four families as $LS\times LS$, $LS \times TS$, $TS \times LS$ and $TS \times TS$.
In the case of an homogeneous undirected graph, only three sets can be defined as the two sets $LS \times TS$ and $TS \times LS$ are confounded. \cite{schrynemackers2013}

Prediction performance are expected to differ between these four families. 
Typically, one expects that $TS \times TS$ pairs will
be the most difficult to predict since less information is available
at training about the corresponding nodes. 
These predictions will then be evaluated separately in this work.

Cross-validation procedures evaluate supervised network inference methods.
A first procedure (cross-validation on pairs) evaluates $LS \times LS$ pairs and is represented in Figure~\ref{protocols2}(B).
A second one (cross-validation on nodes) evaluates $LS \times TS$, $TS \times LS$ and $TS \times TS$ pairs and is represented in Figure~\ref{protocols2}(C). \cite{schrynemackers2013}


\section{Methods}

\label{methods}

In this section, we first present the two generic, local and global, approaches
we have adopted for dealing with classification on pairs. We then discuss their
practical implementation in the context of tree-based ensemble methods. In the
presentation of the approaches, we will assume that we have at our disposal a
classification method that derives its classification model from a class
conditional probability model. Denoting by $f:{\mathcal X}\rightarrow \{0,1\}$ a
classification model (defined on some input space ${\mathcal X}$), we will denote by
$f^p:{\mathcal X}\rightarrow [0,1]$ (i.e., with superscript $p$) the corresponding
class conditional probability function (with $f(x)=1(f^p(x)>p_{th})$ for some
user-defined threshold $p_{th}\in [0,1]$).


\subsection{Global Approach}

\begin{figure}
\centering
\includegraphics[width=9.5cm]{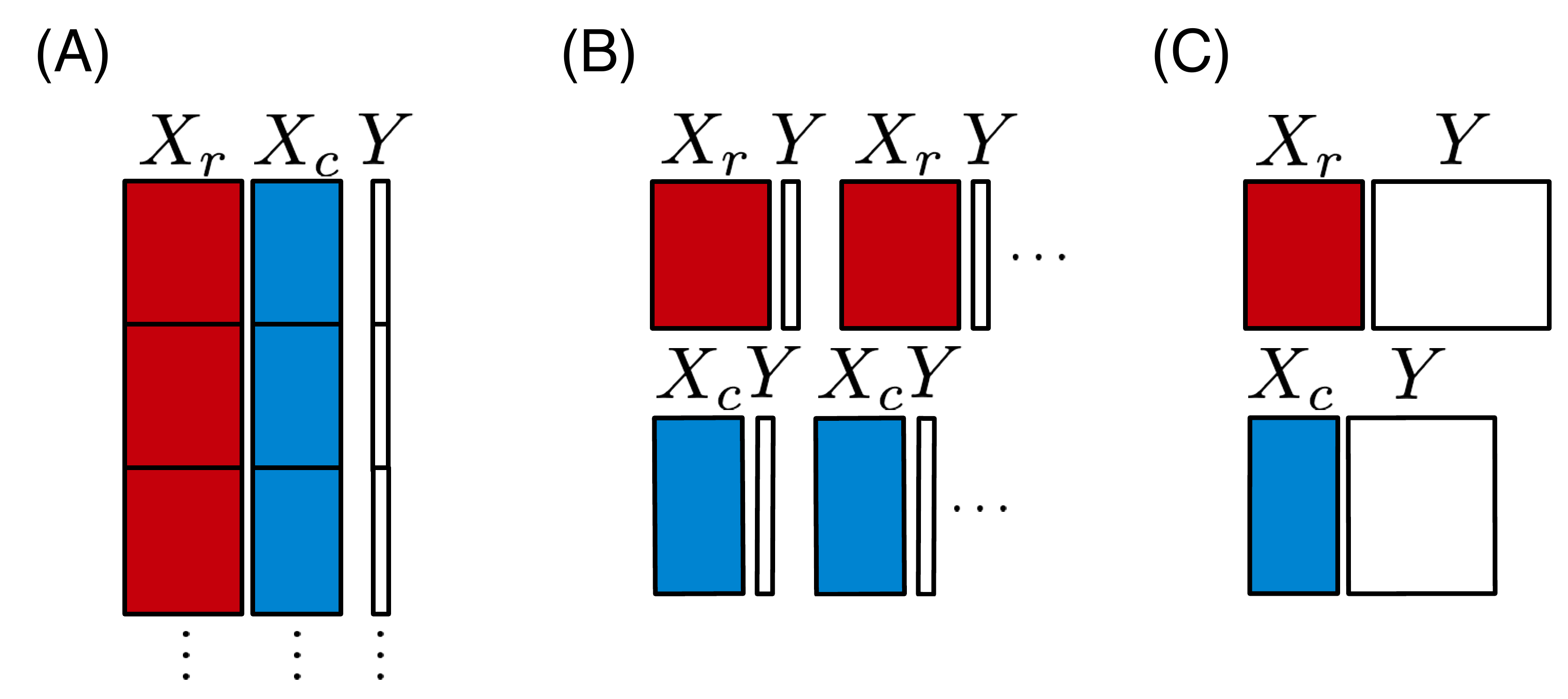}
\caption{Schematic representation of the training data. In the global approach (A) the features vectors are concatenated, in the local approach with single output (B) one function is learnt for each node, and in the local approach with multiple output (C) one function is learnt for one family of nodes and one function for the other one.}
\label{glob-loc-locmo}
\end{figure}

The most straightforward approach for dealing with the problem defined
in Section~\ref{sec:classpair} is to apply a classification algorithm
on the learning sample $LS_p$ of pairs to learn a function
$f_{glob}:U_r\times U_c\rightarrow \{0,1\}$
on the cartesian product of the two input spaces (resulting in
the concatenation of the two input vectors of the nodes of the
pair). Predictions can then be computed straightforwardly for
any new unseen pair from the function. (Figure~\ref{glob-loc-locmo}(A))

In the case of a homogeneous graph, the output function
$y$ is symmetric, i.e., $y(n_r,n_c)=y(n_c,n_r)$, $\forall n_r,n_c \in U$. 
We will introduce two adaptations of the approach to handle such
graphs. First, for each pair $(n_r,n_c)$ in the learning sample, the pair
$(n_c,n_r)$ will also be introduced in the learning sample. Without further
constraint on the classification method, this will not ensure however that the
learnt function $f_{glob}$ will be symmetric in its arguments. To make it symmetric, we will
compute a new class conditional probability model $f^p_{glob,sym}$ from the
learned one $f^p_{glob}$ as follows:
$$f^p_{glob,sym}(x_1,x_2)=\frac{f^p_{glob}(x_1,x_2)+f^p_{glob}(x_2,x_1)}{2}.$$

\subsection{Local Approach}

The idea of the local approach as first proposed in \cite{bleakley2007}, is to build a
separate classification model for each node, trying to predict its neighbors
in the graph from the known graph around this node. More precisely,
for every node $n_c\in LS_c$, a new learning sample $LS(n_c)$ is
constructed from the learning sample of pairs $LS_p$ as follows:
$$
LS(n_c)=\{\langle n_r,y(n_r,n_c)\rangle | \langle n_r,n_c,y(n_r,n_c)\rangle \in LS_p\}.
$$

It can then be used to learn a
classification model $f_{n_c}:U_r\rightarrow \{0,1\}$, which can be
exploited to make a prediction for any new pair $(n'_r,n'_c)$ such that $n'_c=n_c$. By symmetry, the same strategy
can be adopted to learn a classification model $f_{n_r}:U_c\rightarrow
\{0,1\}$ for each node $n_r \in LS_r$. (Figure~\ref{glob-loc-locmo}(B))

These two sets of classifiers can then be exploited to make $LS \times TS$
and $TS \times LS$ types of predictions. For pairs $(n_r,n_c)$ in $LS_r
\times LS_c\setminus LS_p$, two predictions can be obtained: $f_{n_c}(n_r)$
and $f_{n_r}(n_c)$. We propose to simply combine them by an arithmetic
average of the corresponding class conditional probability estimates:
$$f^p_{loc}(n_r,n_c)=\frac{f^p_{n_c}(n_r)+f^p_{n_r}(n_c)}{2}.$$

\begin{figure}
\begin{center}
\includegraphics[width=7cm]{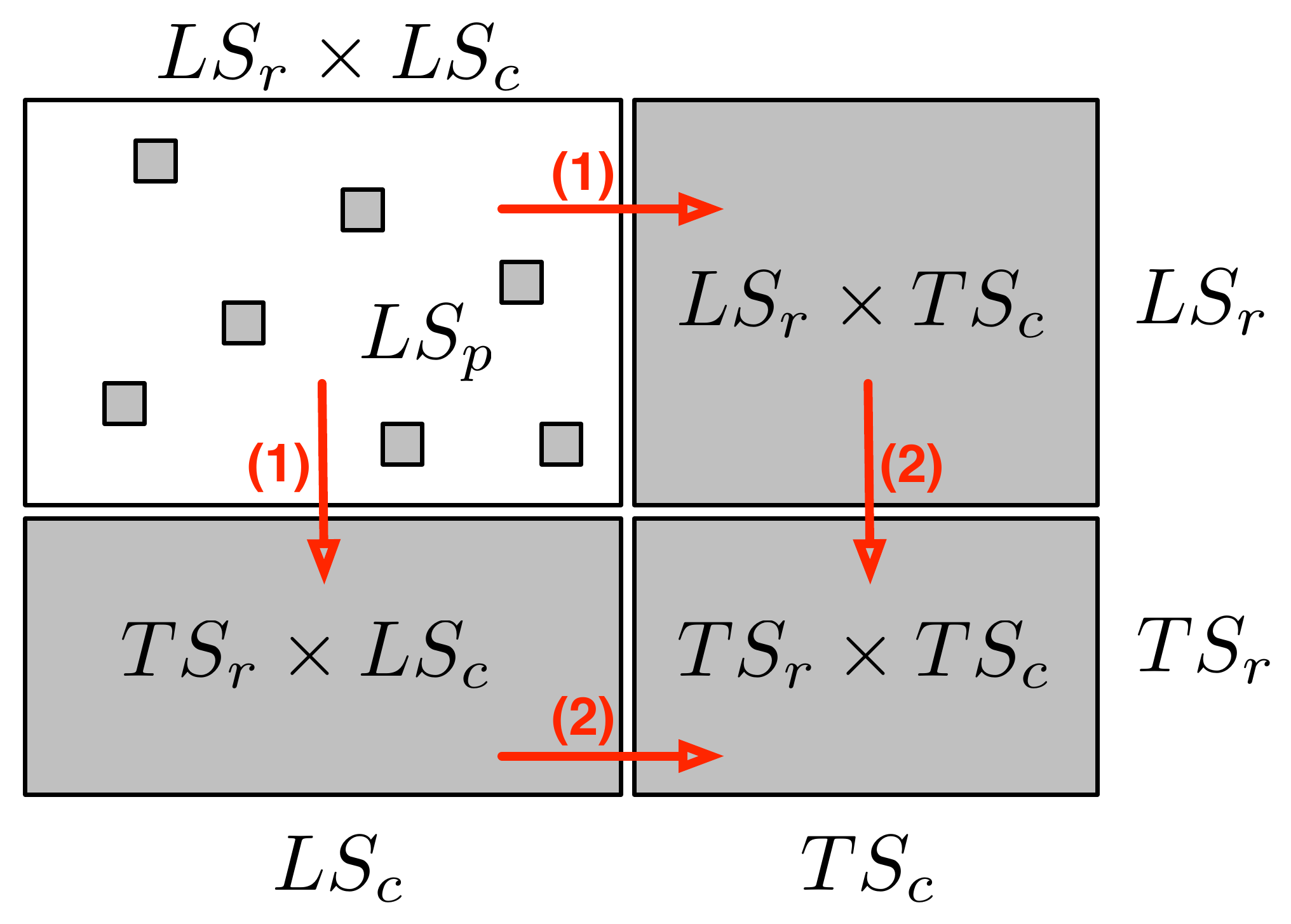}
\caption{The local approach needs two steps to learn a classifier for an unseen node: first, we predict $LS\times TS$ and $TS \times LS$ interactions, and from these predictions, we predict $TS \times TS$ interactions.}
\label{fig:localtsts}
\end{center}
\end{figure}

As such, the local approach is in principle not able to make directly
predictions for pairs of nodes $(n_r,n_c) \in TS \times TS$
(because $LS(n_r)=LS(n_c)=\emptyset$ for $n_r\in TS_r$ and $n_c\in
TS_c$). We nevertheless propose to use the following two-steps
procedure to learn a classifier for a node $n_r \in TS_r$ (see Figure \ref{fig:localtsts}):
\begin{itemize}
\item First, learn all classifiers $f_{n_c}$ for nodes $n_c \in LS_c$, 
\item Then, learn a classifier $f^f_{n_r}:U_c\rightarrow \{0,1\}$ from
$LS^f(n_r)=\{\langle n_c,f_{n_c}(n_r)\rangle | n_c \in LS_c\}$,
i.e., the predictions given by the models $f_{n_c}$ trained in the
first step. 
\end{itemize}
Again by symmetry, the same strategy can be applied to
obtain models $f^f_{n_c}$ for the nodes $n_c \in TS_c$. A prediction
is then obtained for a pair $(n_r,n_c)$ in $TS \times
TS$ by averaging the class conditional probability predictions of
both models $f^{f,p}_{n_r}$ and $f^{f,p}_{n_c}$:
$$
f^p_{loc}(n_r,n_c)=\frac{f^{f,p}_{n_r}(n_c)+f^{f,p}_{n_c}(n_r)}{2}.
$$ Besides
averaging, we tried several alternative schemes to merge the two
models (such as taking the min, max, or the product of their
predictions) but they did not lead to any improvement. Note that
building the learning samples $LS^f(n_r)$ and $LS^f(n_c)$ requires to
choose a threshold on the class conditional probability estimates. In
our experiments, we will set this threshold in such a way that the
proportion of edges versus non edges in the predicted subnetworks in
$LS \times TS$ and $TS\times LS$ is equal to the same
proportion within the original learning sample of pairs.

This strategy can be specialized to the case of a homogeneous graph in a straightforward way. Only one class of
classifiers $f_n:U\rightarrow \{0,1\}$ and $f^f_n:U\rightarrow \{0,1\}$ are trained for nodes in $LS$ and in
$TS$ respectively (using the same two-step procedure as in the
asymmetric case for the second). $LS \times LS$ and $TS \times
TS$ predictions are still obtained by averaging two predictions, one
for each node of the pair.

\subsection{Tree-based ensemble methods}

Any method could be used as a base classifier for the two approaches. In this
paper, we propose to evaluate the use of tree-based ensemble methods in this
context. We first briefly describe these methods and then discuss several
aspects related to their use within the two generic approaches.

\paragraph{Description of the methods.} 

A decision tree \cite{breiman1984} represents an input-output model by a
tree whose interior nodes are each labeled with a (typically binary)
test based on one input feature and each terminal node is labeled with
a value of the output. The predicted output for a new instance is
determined as the output associated to the leaf reached by the
instance when it is propagated through the tree starting at the root
node. A tree is built from a learning sample of input-output pairs, by
recursively identifying, at each node, the test that leads to a split of
the nodes sample into two subsamples that are as pure as possible in
terms of their output values.

Single decision trees typically suffer from high variance, which makes
them not competitive in terms of accuracy. This problem is
circumvented by using ensemble methods that generate several trees and
then aggregate their predictions. In this paper, we exploit one
particular ensemble method called extremely randomized trees
(Extra-trees, \cite{geurts2006et}). This method grows each tree in the
ensemble by selecting at each node the best among $K$ randomly
generated splits. In our experiments, we use the default setting of
$K$, equal to the square root of the total number of candidate
attributes.

One interesting features of tree-based methods (single and ensemble)
is that they can be extented to predict a vectorial output instead of
a single scalar output \cite{blockeel-1998}. We will exploit this
feature of the method in the context of the local approach below.





\begin{figure}
\centerline{\includegraphics[width=9cm]{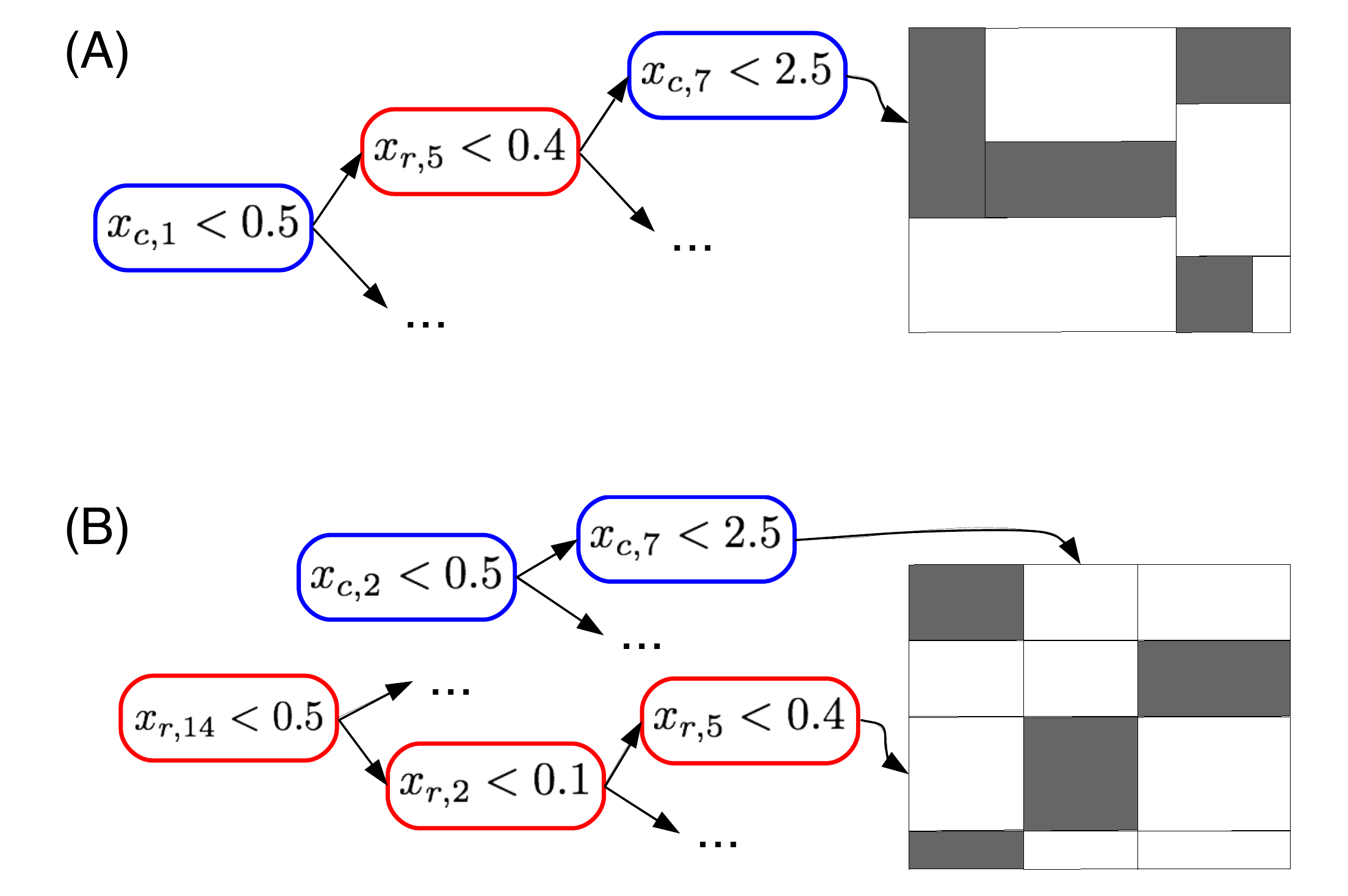}}
\caption{Both the global approach (A) and the local approach with
  multiple output (B) can be interpreted as carrying out a
  biclustering of the adjacency matrix. Note that in the case of the
  global approach, the representation is only illustrative. The
  adjacency submatrices corresponding to the tree leaves can not be
  necessarily rearranged as contiguous rectangular submatrices
  covering the initial adjacency matrix.}
\label{interp_asym}
\end{figure}

\paragraph{Global approach.} The global approach consists in building a tree
from the learning sample of all pairs. Each split of the resulting
tree will be based on one of the input features coming from either one
of the two input feature vectors, $x(n_r)$ or $x(n_c)$. The tree
growing procedure can thus be interpreted as interleaving the
construction of two trees: one on the row nodes and one on the
column nodes. Each leaf of the resulting tree is thus associated
with a rectangular submatrix of the graph adjacency matrix
$Y(LS_r,LS_c)$ and the construction of the tree is such that the pairs
in this submatrix should be, as far as possible, either all connected
or all disconnected (see Figure \ref{interp_asym} for an
illustration).

\paragraph{Local approach.} The use of tree ensembles in the context of the
local approach is straightforward. We will nevertheless compare two
variants. The first one builds a separate model for each row and column nodes
as presented in Section \ref{methods}. The second method exploits the ability
of tree-based methods to deal with multiple outputs to build only two models,
one for the row nodes and one for the column nodes (Figure \ref{glob-loc-locmo}(C)). Assuming that the
learning sample has been generated by sampling two subsets of objets
$LS_r=\{n^1_r,\ldots,n^{N_r}_r\}$ and $LS_c=\{n^1_c,\ldots,n^{N_c}_c\}$ and
that the full adjacency matrix is observed between these two sets, these two
models are built from the following learning samples:
\begin{eqnarray*}
LS(n_c) & = & \{\langle n^i_r,(y(n^i_r,n^1_c),\ldots,y(n^i_r,n^{N_c}_c))\rangle | i=1,\ldots,N_r\},\\
LS(n_r) & = & \{\langle n^j_c,(y(n^1_r,n^j_c),\ldots,y(n^{N_r}_r,n^j_c))\rangle | j=1,\ldots,N_c\}.
\end{eqnarray*}
The same multiple output approach can then be applied to build the two models
required to make $TS\times TS$ predictions. This approach has the advantage
of requiring only four tree ensemble models in total instead of
$N^U_r+ N^U_c$ models for the single output approach. It
can however only be used when the complete submatrix is observed for pairs in
$LS\times LS$, since tree-based ensemble method can not cope with missing
output values.

\paragraph{Interpretability.} 

One main advantage of tree-based methods is their interpretability,
directly through the tree structure in the case of single tree models
and through feature importance rankings in the case of ensembles
\cite{geurts2009}. Let us to compare both approaches along this
criterion.

In the case of the global approach, as illustrated in
Figure~\ref{interp_asym}(A), the tree that is built partitions the
adjacency matrix $Y(LS_r,LS_c)$ into rectangular regions. These regions are defined such that
pairs in each region are either all connected or all disconnect. The
region is furthermore characterized by a path in the tree (from the
root to the leaf) corresponding to tests on the input features of both
nodes of the pair. In the case of the local multiple output
approach, one of the two trees partitions the rows and the other tree
partitions the column of the matrix $Y(LS_r,LS_c)$. Each partitioning
is carried out in such a way that nodes in each subpartition has a
similar connectivity profiles. The resulting partitioning of the
adjacency matrix will thus follow a checkerboard structure with also
only connected or disconnected pairs in the obtained submatrix, as far
as possible (Figure \ref{interp_asym}(B)). Each submatrix will be
furthermore characterized by two conjunctions of tests, one based on
row inputs and one based on column inputs. These two methods can thus
be interpreted as carrying out a biclustering \cite{madeira:2004tj}
of the adjacency matrix where the biclustering is however directed by
the choice of tests on the input features.  These methods are to
biclustering what predictive clustering trees \cite{blockeel-1998}
are to clustering. In the case of the local single output approach,
the partitioning is more fine-grained as it can be different from one
row or column to another. However in this case, each tree gives an
interpretable characterization of the nodes which are connected to
the node from which the tree was built.

When using ensembles, the global approach provides a global ranking of
all features from the most to the less relevant. The local
multiple output approach provides two separate rankings, one for the
row features and one for the column features and the local single
output approach gives a separate ranking for each node. All variants
are therefore complementary from an interpretability point of view.

\paragraph{Implementation and computational issues.}

In principle, since tree building is a batch algorithm, the global
approach requires to generate the full sample of all pairs, which may
be very prohibitive for graphs defined on a large number of nodes
(e.g., the PPI network used in our experiments contains about 1000
nodes that lead to about 1 millions pairs described by 650
attributes). Fortunately, since the tree building method goes through
the input features one by one, one can separately search for the best
split on features relative to nodes in $U_r$ and on features relative to nodes in $U_c$, which does 
not require to generate explicitly the full data
matrix. This is an important advantage with respect to kernel-based
methods that typically requires to handle explicitly a $N_rN_c \times
N_rN_c$ Gram matrix. Since tree growing is order $O(N \log(N))$ for a
training sample of size $N$, the computational complexity of the whole
procedure however remains $O(N_c N_r (\log(N_c)+\log(N_r)))$. The
complexity of the trees (measured by the total number of nodes) is at
worst $O(N_cN_r)$ (corresponding to a fully developed tree) but in
practice it is related to the number of positive interactions in the
training sample, which is typically much lower than $N_cN_r$.

The computational complexity of the local approach is the same as the
computational complexity of the global approach, i.e.
$O(N_cN_r\log(N_r)+N_rN_c\log(N_c))$. Indeed, in the single output
approach, $N_c$ and $N_r$ models need to be constructed respectively
from $N_r$ samples and $N_c$ samples each. In the multiple output
case, only two models are constructed from $N_r$ and $N_c$ samples
respectively, but the multiple output variant needs to go through all
outputs at each tree node, which multiplies complexity by respectively
$N_r$ and $N_c$ for these two models. However, at worst, the complexity of the
model is $O(N_cN_r)$ for the single output approach and $O(N_r+N_c)$
for the multiple output approach, which potentially gives an important
advantage along this criterion for the multiple output method.


\section{Experiments} 
\label{sec:experiments}

In this section, we carried out a large scale empirical evaluation of
the different methods described in Section~\ref{methods} on six real
biological networks, three homogeneous graphs and three bipartite
graphs. Results on four additional (drug-protein) networks can be
found in appendix~\ref{app:4net}. Our goal with these experiments
is to assess the relative performances of the different approaches and
to give an idea of the performance one could expect from these methods
on biological networks of different nature. Section \ref{sec:comp}
provides a comparison with existing methods from the literature.

\subsection{Datasets}
\label{sec:datasets}

\begin{table}
\caption{Summary of the six datasets used in the experiments.}
\label{tab:datasets}
\vspace{0.3cm}
\begin{tabular}{rllll}
&Network & Network size & Number of edges & Number of features\\
\cline{2-5}
\textit{Homogeneous networks}&PPI & 984$\times$984 & 2438 & 325\\ 
&EMAP & 353$\times$353 & 1995 & 418\\
&MN & 668$\times$668 & 2782 & 325 \vspace{0.1cm} \\
\textit{Bipartite networks}&ERN & 154$\times$1164  &3293 & 445/445\\ 
&SRN & 113$\times$1821 & 3663 & 9884/1685\\
&DPI & 1862$\times$1554 & 4809 & 660/876\\
\cline{2-5}
\end{tabular}
\end{table}

The first three networks correspond to homogeneous undirect graphs and
the last three to bipartite graphs. The main characteristics of the
datasets are summarized in Table \ref{tab:datasets}.

\paragraph{Protein-protein interaction network (PPI).}
This network has been compiled from the 2438 high confidence
interactions between 984 {\it S.cerivisiae} proteins highlighted by
\cite{vonmering2002}. The input features used for the predictions are a set of
expression data, phylogenetic profiles and localization data that
totalizes 325 features. This dataset has been used in several studies
before \cite{yamanishi04,kato2005,geurts2007}.

\paragraph{Genetic interaction network (EMAP).}
This network \cite{schuldiner2005} contains 353 {\it S.cerivisiae} genes
(nodes) connected with 1995 negative epistatic interactions
(edges). Inputs consists in measures of growth fitness of yeast celln relative to deletion of each gene separately, and in 418 different environments. \cite{hillenmeyer2008}.

\paragraph{Metabolic network (MN).}
This network \cite{yamanishi2005} is composed of 668 {\it S.cerivisiae}
enzymes (nodes) connected by 2782 edges. There is an edge between two
enzymes when these two enzymes catalyse successive reactions. The
input feature vectors are the same as those used in the PPI network.

\paragraph{\textit{E.coli} regulatory network (ERN).}
This bipartite graph \cite{faith:2007p3083} connects transcription
factors (TF) and genes of {\it E.coli}. It is composed of 1164 genes
regulated by 154 TF. There is a total of 3293 interactions. The input
features \cite{faith:2007p3083} are 445 expression values.

\paragraph{\textit{S.cerevisiae} regulatory network (SRN).}
This network \cite{macisaac2006} connects TFs and their target genes
from \textit{E.coli}. It is composed of 1855 genes regulated by 113
TFs and totalizing 3737 interactions. The input features are 1685
expression values \cite{hughes2000,hu2007,chua2006,faith2007}. For genes, we
concatenated motifs features \cite{brohee2008} to the expression values.

\paragraph{Drug-protein interaction network (DPI).}
Drug-target interactions \cite{yamanishi:2011ca} are related to
human and connect a drug with a protein when the drug targets the
protein. This network holds 4809 interactions involving 1554
proteins and 1862 drugs. The input features are a binary vectors coding for
the presence or absence of 660 chemical substructures for each drug,
and the presence or absence of 876 PFAM domains for each protein
\cite{yamanishi:2011ca}.

\subsection{Protocol}
\label{sec:protocol}

In our experiments, $LS\times LS$ performances in each network are
measured by 10 fold cross-validation (CV) across the pairs of nodes, as illustrated in Figure~\ref{protocols2}(B).
For robustness, results are averaged over 10
runs of 10 fold CV. $LS \times TS$, $TS \times LS$ and $TS \times TS$ predictions are
assessed by performing a 10 times 10 fold CV across the nodes, as illustrated in Figure~\ref{protocols2}(C). 
The different algorithms return class conditional probability
estimates. To assess our models independently of a particular choice of
discretization threshold $P_{th}$ on these estimates, we vary this threshold
and output in each case the resulting precision-recall curve and the
resulting ROC curve. Methods are then compared according to the total
area under these curves, denoted AUPR and AUROC respectively (the
higher the AUPR and the AUROC, the better), averaged over the 10 folds
and the 10 CV runs. For all our experiments, we use ensembles of 100
extremely randomized trees with default parameter setting
\cite{geurts2006et}.

As highlighted by several studies, e.g. \cite{gillis2011}, in
biological networks, nodes of high degree have a higher chance to be
connected to any new node. In our context, this means that we can
expect that the degree of a node will be a good predictor to infer new
interactions involving this node. We want to assess the importance of this effect and provide a more realistic
baseline than the usual random guess performance. To reach this goal, we
evaluate the AUROC and AUPR scores when using the sum of the
degrees of each node in a pair to rank $LS\times LS$ pairs and when
using the degree of the nodes belonging to the LS to rank $TS\times LS$
or $LS\times TS$ pairs. AUROC and AUPR scores will be evaluated using
the same protocol as hereabove. As there is no information about the
degrees of nodes in $TS\times TS$ pairs, we will use random guessing
as a baseline for the scores of these predictions (corresponding to an
AUROC of 0.5 and an AUPR equal to the proportion of interactions among
all nodes pairs).

\subsection{Results}

We discuss successively the results  on the three homogeneous
graphs and then on the three bipartite graphs.

\paragraph{Homogeneous graphs.}

\begin{figure}
\centering
\includegraphics[width=5cm]{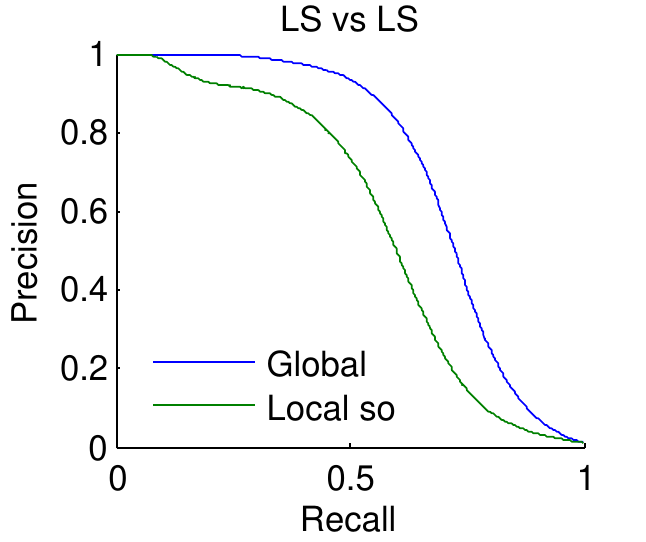}\includegraphics[width=5cm]{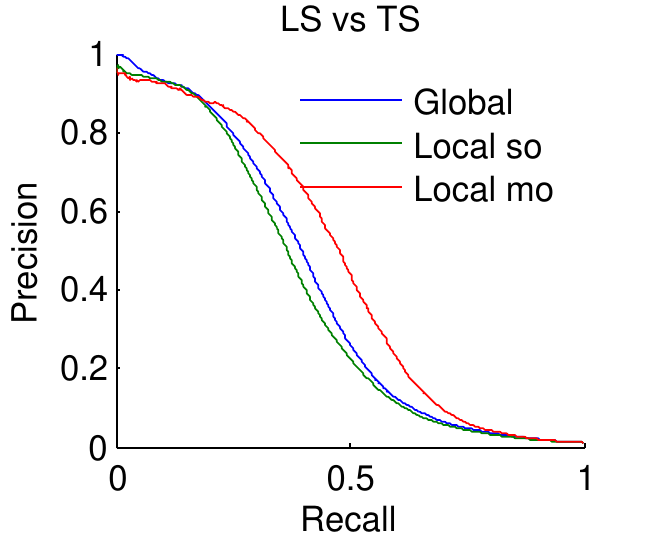} \includegraphics[width=5cm]{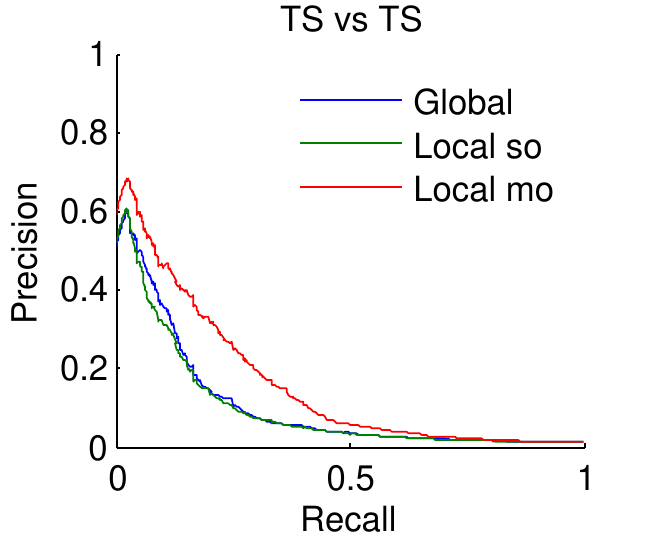}
\caption{Precision-recall curves for metabolic network: higher is the number of nodes of a pair present in the learning set, better will be the prediction for this pair.}
\label{mn}
\end{figure}

AUPR and AUROC values are summarized in Table \ref{auc_ppi} for the
three methods: global, local single output, and local multiple
output. The last row on each dataset is the baseline result obtained
as described in \ref{sec:protocol}. Figure~\ref{mn} shows the
precision-recall curves obtained by the different approaches on MN,
for the three different protocols. Similar curves for the two other
networks can be found in appendix \ref{app:homo}.


\begin{table}
\caption{Areas under curves for homogeneous networks.}
\label{auc_ppi}
\vspace{0.3cm}
\centering
\begin{tabular}{rp{2cm}llp{1.5cm}lll}
&&\multicolumn{3}{l}{Precision-Recall (AUPR)}&\multicolumn{3}{l}{ROC (AUC)} \\ 
 &&LS-LS&LS-TS&TS-TS&LS-LS&LS-TS&TS-TS\\ \cline{2-8}
\textit{PPI}&Global&0.41&0.22&0.10&0.88&0.84&0.76\\ 
&Local so&0.28&0.21&0.11&0.85&0.82&0.73\\ 
&Local mo&-&0.22&0.11&-&0.83&0.72\\ 
&Baseline &0.13&0.02&0.00&0.73&0.74&0.50
\vspace{0.2cm} \\
\textit{EMAP}&Global&0.49&0.36&0.23&0.90&0.85&0.78\\ 
&Local so&0.45&0.35&0.24&0.90&0.84&0.79\\ 
&Local mo&-&0.35&0.23&-&0.85&0.80\\ 
&Baseline &0.30&0.13&0.03&0.87&0.80&0.50 
\vspace{0.2cm} \\
\textit{MN}&Global&0.71&0.40&0.09&0.95&0.85&0.69\\ 
&Local so&0.57&0.38&0.09&0.92&0.83&0.68\\ 
&Local mo&-&0.45&0.14&-&0.85&0.71\\ 
&Baseline &0.05&0.04&0.01&0.75&0.70&0.50\\
\cline{2-8}
\end{tabular}
\end{table}

In terms of absolute AUPR and AUROC values, $LS\times LS$ pairs are
clearly the easiest to predict, followed by $LS\times TS$ pairs and
$TS\times TS$ pairs. This ranking was expected from previous
discussions. Baseline results in the case of $LS\times LS$ and
$LS\times TS$ predictions confirm that nodes degrees are very
informative: baseline AUROC values are much greater than 0.5 and
baseline AUPR values are also significantly higher than the proportion
of interactions among all pairs (0.005, 0.03, and 0.01
respectively for PPI, EMAP, and MN), especially in the case of
$LS\times LS$ predictions. Nevertheless, our
methods are better than these baselines in all cases. On the EMAP
network, the difference in terms of AUROC is very slight but the
difference in terms of AUPR is important. This is typical of highly
skewed classification problems, where precision-recall curves are
known to give a more informative picture of the performance of an
algorithm than ROC curves \cite{davis2006}.

All tree-based approaches are very close on $LS\times TS$ and
$TS\times TS$ pairs but the global approach has an advantage over the
local one on $LS\times LS$ pairs. The difference is important on the
PPI and MN networks. For the local approach, the performance of single
and multiple output approaches are indistinguishable, except with the
metabolic network where the multiple output approach gives better
results. This is in line with the better performance of the global
versus the local approach on this problem, as indeed both the global
and the local multiple output approaches grow a single model that can
potentially exploit correlations between the outputs. Notice that the
multiple output approach is not feasible when we want to predict
$LS\times LS$ pairs, as we are not able to deal with missing output
values in multiple output decision trees.


\paragraph{Bipartite graphs.}

\begin{figure}
\centering
\includegraphics[width=5cm]{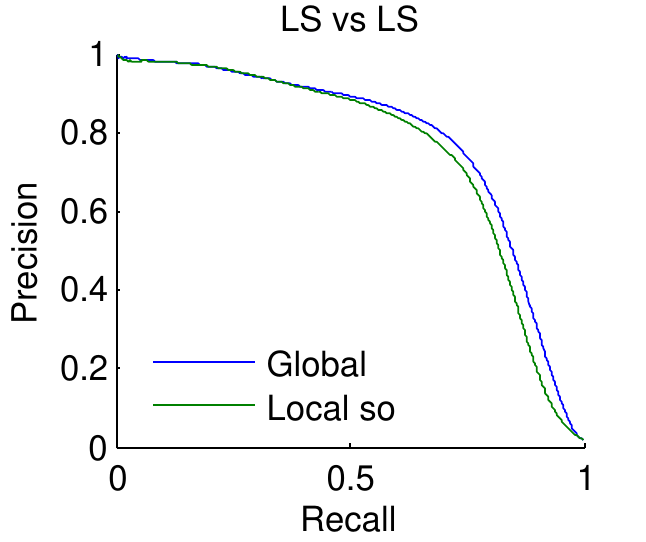}\includegraphics[width=5cm]{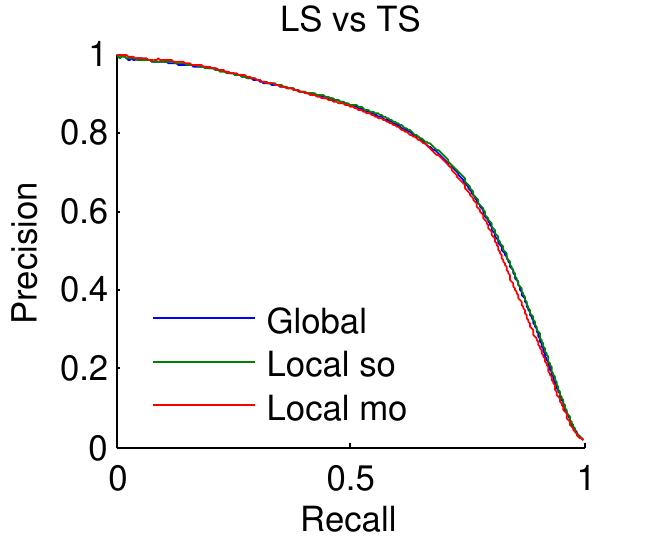}\\
\includegraphics[width=5cm]{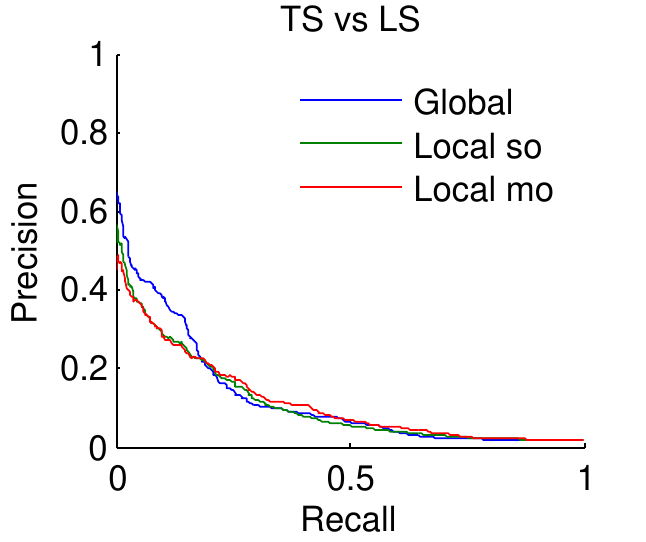}\includegraphics[width=5cm]{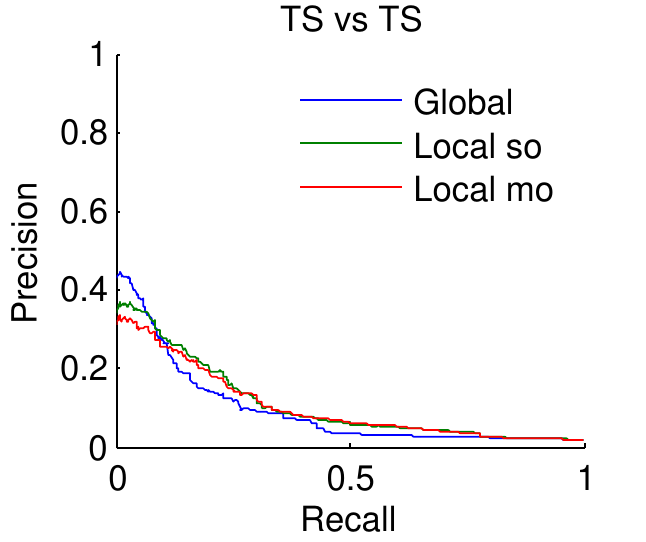}
\caption{Precision-recall curves for E.coli regulatory network (TF vs genes): a prediction is easier to do if the TF belongs to the learning set than if the gene belongs to. \label{ern}}
\end{figure}

AUPR and AUROC values are summarized in Table~\ref{auc_ern} (see
appendix \ref{app:4net} for additional results on four DPI
subnetworks). Figure~\ref{ern} shows the precision-recall curves
obtained by the different approaches on ERN for the four different
protocols. Curves for the 6 other networks can be found in
appendix \ref{app:bip}. 10 times 10-fold CV was used as
explained in Section \ref{sec:protocol}. 
Nevertheless, two difficulties appeared  in the experiments performed on the DPI network. 
First, the dataset is larger than the others, and the 10-fold CV was replaced by 5-fold CV to reduce the computational space et time burden.
Second, the feature vectors are binary and the randomization of the threshold (in Extra-Tree algorithm) cannot lead to diversity between the different trees of the ensemble. So we used bootstrapping to generate the training set of each tree. 

\begin{table}
\caption{Areas under curves for bipartite networks.}
\label{auc_ern}
\vspace{0.3cm}
\centering
\small{
\begin{tabular}{rp{1.7cm}lllp{1.3cm}llll}
&&\multicolumn{4}{l}{Precision-Recall (AUPR)}&\multicolumn{4}{l}{ROC (AUC)} \\
& &LS-LS&LS-TS&TS-LS&TS-TS&LS-LS&LS-TS&TS-LS&TS-TS\\ \cline{2-10}
\textit{ERN} (TF - gene)&Global&0.78&0.76&0.12&0.08&0.97&0.97&0.61&0.64\\ 
&Local so&0.76&0.76&0.11&0.10&0.96&0.97&0.61&0.66\\ 
&Local mo&-&0.75&0.09&0.09&-&0.97&0.61&0.65\\ 
&Baseline&0.31&0.30&0.02&0.02&0.86&0.87&0.52&0.50
\vspace{0.2cm}\\
\textit{SRN} (TF - gene)&Global&0.23&0.27&0.03&0.03&0.84&0.84&0.54&0.57\\ 
&Local so&0.20&0.25&0.02&0.03&0.80&0.83&0.53&0.57\\ 
&Local mo&-&0.24&0.02&0.03&-&0.83&0.53&0.57\\ 
&Baseline&0.06&0.06&0.03&0.02&0.79&0.78&0.51&0.50
\vspace{0.2cm}\\
\textit{DPI} (drug - protein)&Global&0.14&0.05&0.11&0.01&0.76&0.71&0.76&0.67\\ 
&Local so&0.21&0.11&0.08&0.01&0.85&0.72&0.72&0.57\\ 
&Local mo&-&0.10&0.08&0.01&-&0.72&0.71&0.60\\ 
&Baseline&0.02&0.01&0.01&0.01&0.82&0.63&0.68&0.50\\
\cline{2-10}
\end{tabular}}
\end{table}

Like for the homogeneous networks, higher is the number of nodes of
a pair present in the learning set, better are the predictions, ie.,
AUPR and AUROC values are significantly decreasing from $LS\times LS$ to
$TS\times TS$ predictions. On the ERN and SRN networks, performances
are very different for the two kinds of $LS\times TS$ predictions that
can be defined, with much better results when generalizing over genes
(i.e., when the TF of the pair is in the learning sample). On the
other hand, on the DPI network, both kinds of $LS\times TS$
predictions are equally well predicted. These differences are probably
due in part to the relative numbers of nodes of both kinds in the
learning sample, as there are much more genes than TFs on ERN and SRN
and a similar number of drugs and proteins in the DPI
network. Differences are however probably also related to the
intrinsic difficulty of generalizing over each node family, as on
the four additional DPI networks (see appendix~\ref{app:4net}),
generalization over drugs is most of the time better than
generalization over proteins, irrespectively of the relative numbers
of drugs and proteins in the training network.  Results are most of
the time better than the baselines (based on nodes degrees for $LS\times
LS$ and $LS\times TS$ predictions and on random guessing for $TS\times
TS$ predictions). The only exceptions are observed when generalizing
over TFs on SRN and when predicting $TS\times TS$ pairs on SRN and
DPI.

The three approaches are very close to each other. Unlike on
homonegeneous graphs, there is no strong difference between the global
and the local approach on $LS\times LS$ predictions: it is slightly
better in terms of AUPR on ERN and SRN but worse on DPI. The single
and multiple output approaches are also very close, both in terms of
AUPR and AUROC. Similar results are observed on the four additional DPI
networks (appendix~\ref{app:4net}).

\subsection{Comparison with related works}\label{sec:comp}

In this section, we compare our methods with several other network
inference methods from the literature. To ensure a fair comparison and
avoid errors related to the reimplementation and tuning of each of
these methods, we choose to rerun our algorithms in similar settings
as in related papers. All comparison results are summarized in Table
\ref{tab:comparison} and discussed below.

\begin{table}
\caption{Comparison with related works on the different networks.}
\label{tab:comparison}
\vspace{0.3cm}
\centering
\begin{tabular}{llllll} 
Publication&DB&Protocol&Measures&Their results&Our results\\ \hline
\cite{bleakley2007}&PPI&$LS \times TS$, 5CV&AUPR&0.25&0.21\\
&MN&&&0.41&0.43\vspace{0.1cm} \vspace{0.2cm} \\
\cite{geurts2007}&PPI&$LS \times TS$, 10CV&AUPR / ROC&0.18 / 0.91&0.22 / 0.84\\
&&$TS \times TS$&&0.09 / 0.86&0.10 / 0.76\\
&MN&$LS \times TS$&&0.18 / 0.85&0.45 / 0.85\\
&& $TS \times TS$&&0.07 / 0.72&0.14 / 0.71\vspace{0.2cm} \\
\cite{mordelet2008}&ERN&$LS \times TS$, 3CV&Recall 60 / 80& 0.44 / 0.18& 0.38 / 0.15\vspace{0.2cm} \\
\cite{yamanishi:2011ca}&DPI&$LS \times LS$, 5CV&AUROC&0.75&0.88\vspace{0.1cm} \vspace{0.2cm} \\
\cite{yamanishi2012}&DPI&$LS \times LS$, 5CV&AUROC&0.87&0.88\\
&&\multicolumn{2}{l}{$LS \times TS$ \& $TS \times LS$}&0.74&0.74\\ \hline
\end{tabular}
\end{table}

\paragraph{Homogeneous graphs.}

\cite{bleakley2007} developed and applied the local approach with support
vector machines to predict the PPI and MN networks and show that it
was superior to several previous works
\cite{yamanishi04,kato2005}. They only consider $LS\times TS$
predictions and used 5-fold CV. Although they exploited
yeast-two-hybrid data as additional features for the prediction of the
PPI network, we obtain very similar performances with the local
multiple output approach (see Table
\ref{tab:comparison}). \cite{geurts2007} use ensembles of output
kernel trees to infer the MN and PPI networks with the same input data
as \cite{bleakley2007}. With the global approach, we obtain similar or
inferior results as \cite{geurts2007} in terms of AUROC but much
better results in terms of AUPR, especially on the MN data.

\paragraph{Bipartite graphs.}

\cite{mordelet2008} use SVM to predict ERN with the local approach, focusing
on the prediction of interactions between known TFs and new genes
($LS\times TS$). They evaluated their performances by the precision at
60\% and 80\% recall respectively, estimated by 3-fold CV (ensuring that
all genes belonging to a same operon are always in the same fold). Our
results with the same protocol (and the local multiple output variant)
are very close although slightly less good. The DPI network was
predicted in \cite{yamanishi:2011ca} using sparse canonical
correspondence analyze (SCCA) and in \cite{yamanishi2012} with the
global approach and $L_1$ regularized linear classifiers using as
input features all possible products of one drug feature and one
protein feature. Only $LS\times LS$ predictions are considered in
\cite{yamanishi:2011ca}, while \cite{yamanishi2012} differentiate
``pair-wise CV'' (our $LS\times LS$ predictions) and ``block-wise CV''
(our $LS\times TS$ and $TS\times LS$ predictions). As shown in Table
\ref{tab:comparison}, we obtain better results than
\cite{yamanishi:2011ca} and similar results as in
\cite{yamanishi2012}. Additional comparisons are presented in
appendix \ref{app:4net} on the four DPI subnetworks.

Globally, these comparisons show that tree-based methods are
competitive on all six networks. Moreover, it has to be noticed that
(1) no other method has been tested over all these problems, and (2)
we have not tuned any parameters of the Extra-trees method. Better
performances could be achieved by changing, for example, the
randomization scheme \cite{breiman2001}, the feature selection parameter $K$,
or the number of trees.

\section{Discussion} \label{concl}


We explored tree-based ensemble methods for biological network
inference, both with the local approach, which trains a
separate model for each network node (single output) or each node
family (multiple output), and with the global approach, which trains a
single model over pairs of nodes. We carried out experiments on ten
biological networks and compared our results with those from the
literature. These experiments show that the resulting methods are
competitive with the state of the art in terms of predictive
performance. Other intrinsic advantages of tree-based approaches
include their interpretability, through single tree structure and
ensemble-derived feature importance scores, as well as their almost
parameter free nature and their reasonable computational complexity
and storage requirement.


While the local and global approaches are close in terms of accuracy,
the most appealing approach in our experiments turns out to be the
local multiple output method, which provides less complex models and
requires less memory at training time. All approaches remain however
interesting because of their complementarity in terms of
interpretability. A potential advantage of the global approach that
was not explored in this paper is the possibility to define features
on pairs of nodes that might make a difference in some applications
\cite{lin2004,qi2005,yamanishi2012}. With the introduction of such
features, one would loose however the possibility with tree-based
methods of not generating explicitely all pairs when training the
model.


As two side contributions, we extended the local approach for the
prediction of edges between two unseen nodes and proposed the use of multiple
output models in this context. The two-step procedure used to obtain
this kind of predictions provides similar results as the global
approach, although it trains the second model on the first model's
predictions. It would be interesting to investigate other
prediction schemes and evaluate this approach in combination with
other supervised learning methods such as SVMs. The main benefits of using multiple
output models is to reduce model sizes and potentially computing
times, as well as to reduce variance, and therefore improving
accuracy, by exploiting potential correlations between the outputs. It
would be interesting to apply other multiple output or multi-label SL
methods \cite{tsoumakas2007multi} within the local approach.


We focused on the evaluation and comparison of our methods on various
biological networks. To the best of our knowledge, no other study has
considered simultaneously as many of these networks. Our protocol
defines an experimental testbed to evaluate new supervised network
inference methods. Given our methodological focus, we have not tried
to obtain the best possible predictions on each and every one of these
networks. Obviously, better performances could be obtained in each
case by using up-to-date training networks, by incorporating other
feature sets, and by (cautiously) tuning the main parameters of
tree-based ensemble methods. Such adaptation and tuning would not
change however our main conclusions about relative comparisons between
methods.



Our experiments, like others \cite{yamanishi2012}, show that the
different families of predictions that are defined by the two protocols
are not equally well predicted, which justifies their separate
assessment.  These discrepancies in terms of prediction quality should
be taken into account when one wants to merge the different families
of pairs into a single ranked list of novel candidate interactions
from the more to the less confident as predicted by our models. This
question largely deserves further study. A limitation of our protocol
is that it assumes the presence of known positive and negative
interactions. Most often in biological networks, only positive
interactions are recorded, while all unlabeled interactions are not
necessarily true negatives (a notable exception in our experiments is
the EMAP dataset). In this work, we considered that all unlabeled
examples are negative examples. It was shown empirically and
theoretically that this approach is reasonable
\cite{elkan08}. It would be interesting nevertheless to
design tree-based ensemble methods that explicitely takes into account
the absence of true negative examples (e.g., \cite{denis:2005p2650}).



\section*{Acknowledgements}

The authors thank the GIGA Bioinformatics platform and the SEGI for
providing computing resources.

\appendix
\section{Appendix}

\subsection{Homogeneous graphs}
\label{app:homo}


\subsubsection{PPI network}
\begin{center}
\includegraphics[width=5cm]{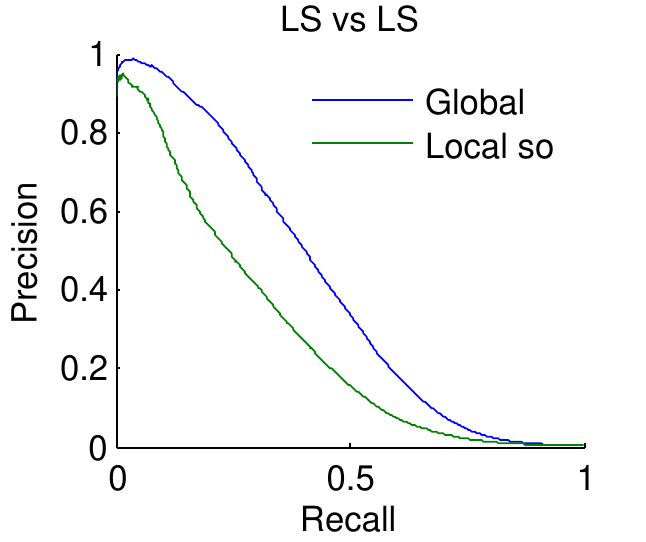}\includegraphics[width=5cm]{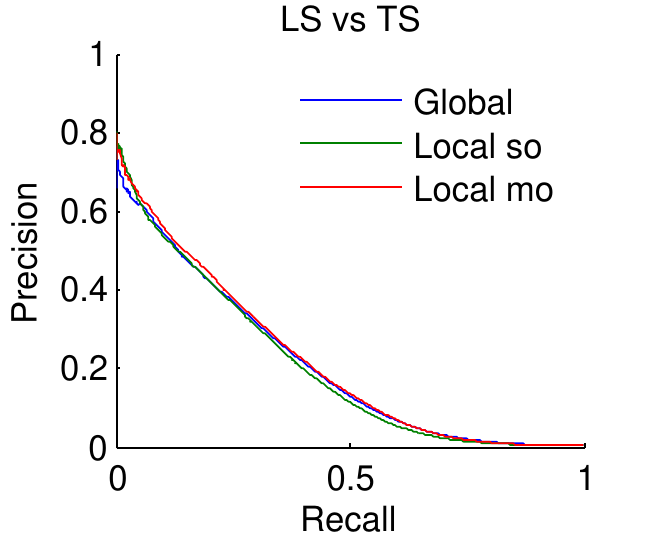}\includegraphics[width=5cm]{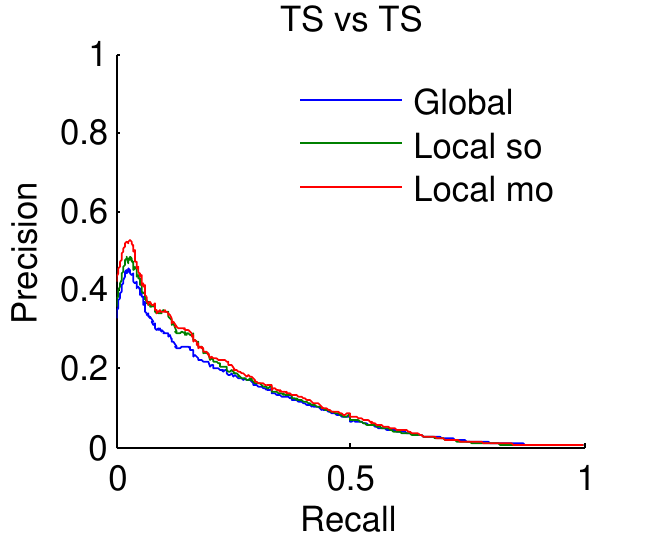}
\end{center}

\subsubsection{EMAP network}
\begin{center}
\includegraphics[width=5cm]{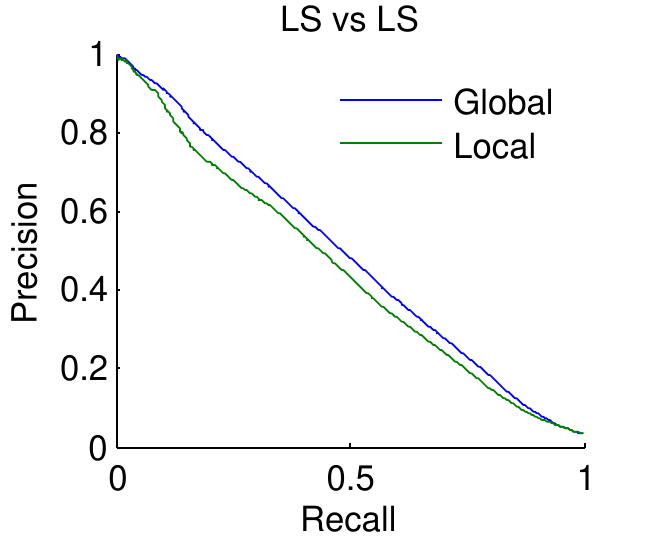}\includegraphics[width=5cm]{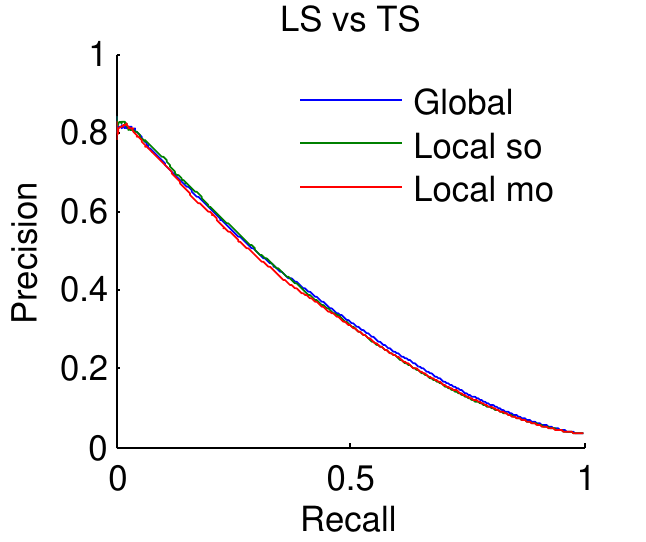}\includegraphics[width=5cm]{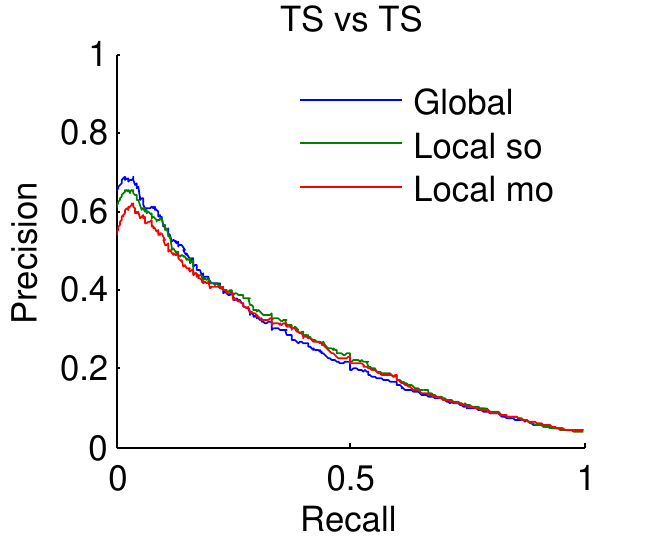}
\end{center}

\subsection{Bipartite graphs}
\label{app:bip}


\subsubsection{S.cerevisiae regulatory network (TF vs genes)}
\begin{center}
\includegraphics[width=5cm]{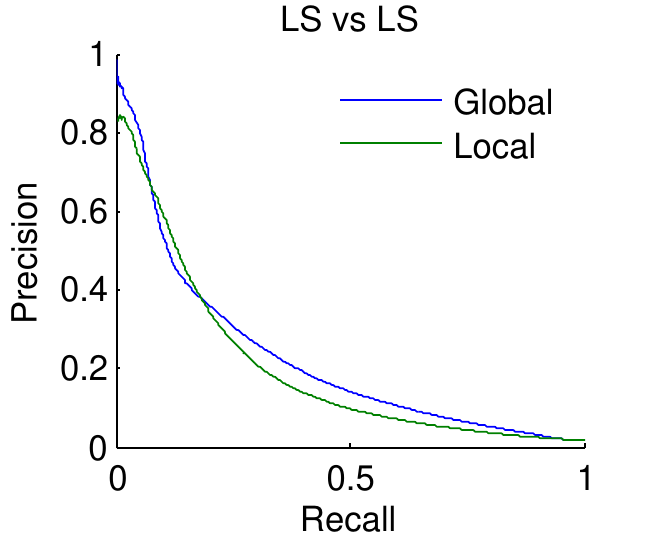}\includegraphics[width=5cm]{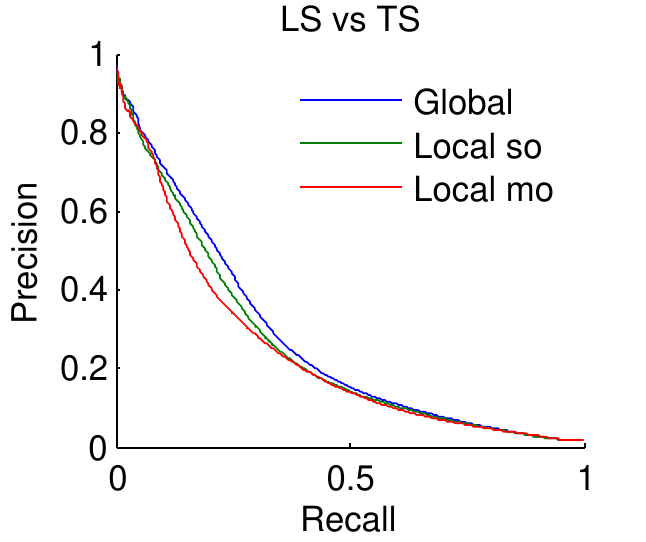} \\
\includegraphics[width=5cm]{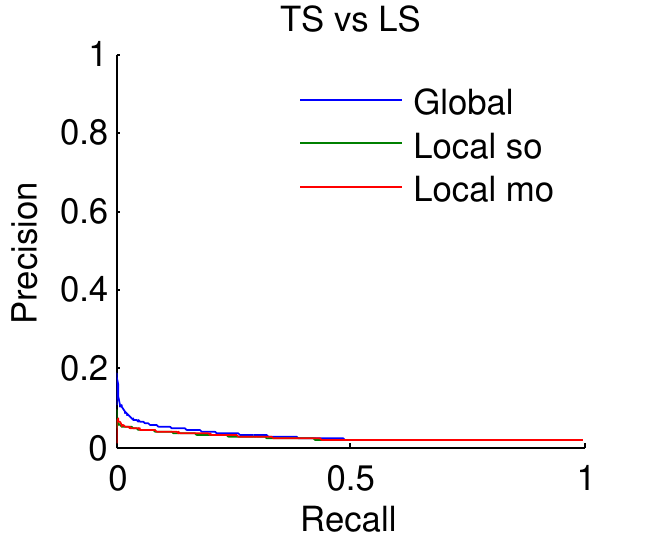}\includegraphics[width=5cm]{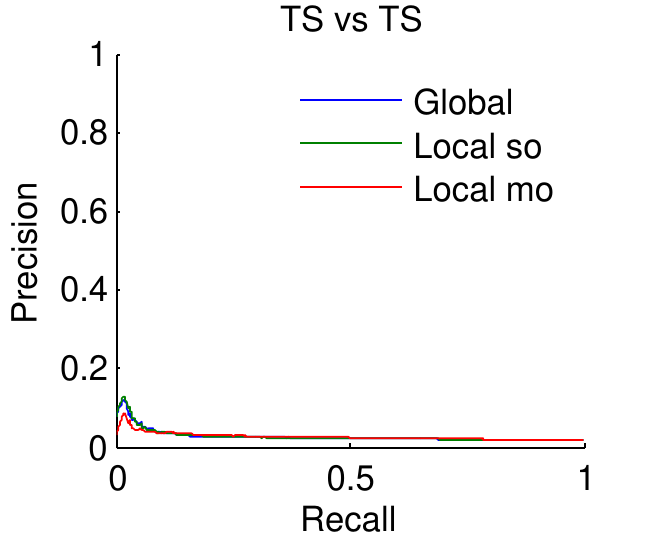}
\end{center}

\subsubsection{Drug-protein interaction network}
\begin{center}
\includegraphics[width=5cm]{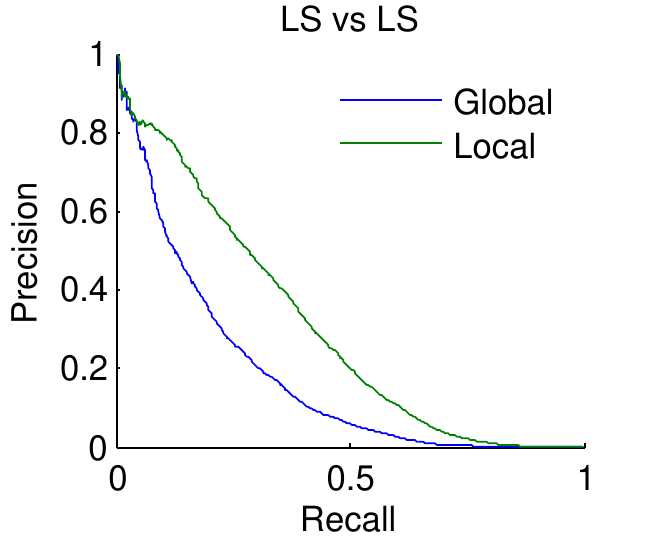}\includegraphics[width=5cm]{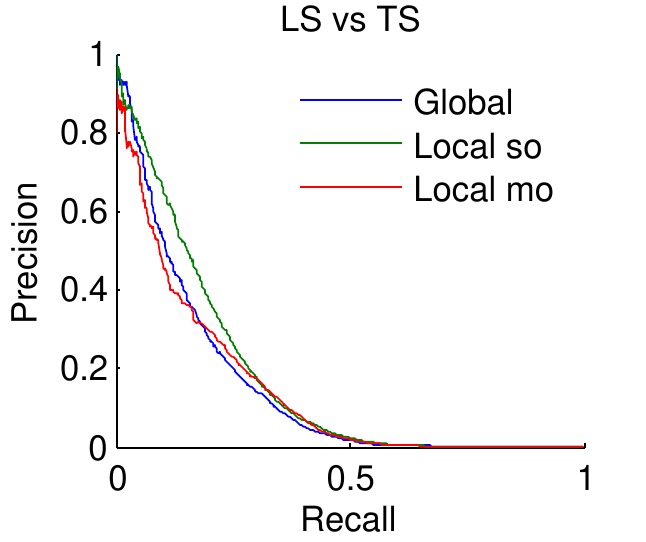} \\
\includegraphics[width=5cm]{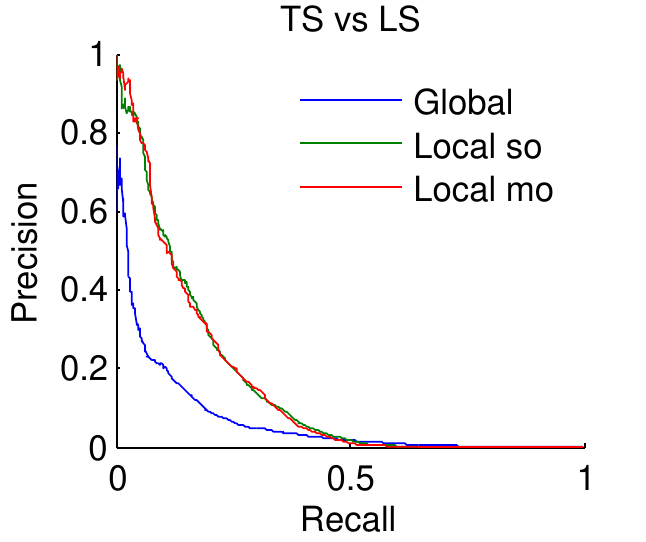}\includegraphics[width=5cm]{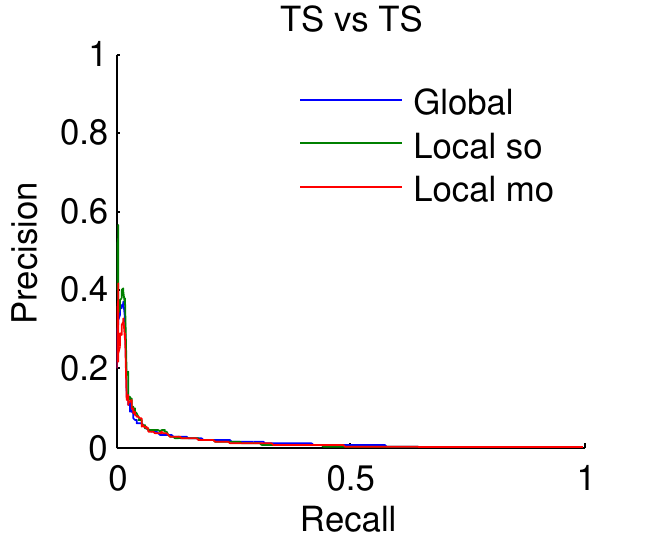}
\end{center}

\subsubsection{Four kinds of drug-protein interaction networks}
\label{app:4net}

\paragraph{Datasets.}

\cite{yamanishi2008} proposed four different drug-protein interaction networks in which proteins belong to four pharmaceutically useful classes: enzymes (DPI-E), ion channels (DPI-I), G-protein-coupled receptors (DPI-G) and nuclear receptors (DPI-N). The input features for proteins are similarity with all proteins in terms of sequence and the input features for drugs are similarity with all drugs in terms of chemical structure \cite{yamanishi2008}. The number of drugs in these networks are respectively 445, 210, 223 and 54, the number of proteins are 664, 204, 95 and 26 and the number of interactions are 2926, 1476, 635 and 90.\\

\begin{center}
\begin{tabular}{p{1.5cm}p{2.1cm}p{1.5cm}p{2.5cm}}
Network & Network size & \# edges & \# input features\\
\hline
DPI-E & 445$\times$664 & 2926 & 445/664\\
DPI-I & 210$\times$204 & 1476 & 210/204\\
DPI-G & 223$\times$95 & 635 & 223/95\\
DPI-N & 54$\times$26 & 90 & 54/26\\
\hline
\end{tabular}
\end{center}

\paragraph{Results.}
Areas under precision-recall and ROC curves for the four networks:

\begin{center} 
\begin{tabular}{p{1.8cm}p{1.1cm}p{1.1cm}p{1.1cm}p{1.1cm}p{1.1cm}p{1.1cm}p{1.1cm}p{1.1cm}}
&\multicolumn{4}{l}{Precision-Recall}&\multicolumn{4}{l}{ROC} \\ \hline
&LS-LS&LS-TS&TS-LS&TS-TS&LS-LS&LS-TS&TS-LS&TS-TS\\ \hline
\multicolumn{9}{l}{\textit{Drug-protein (enzyme) interaction network}}\\
Global&0.86&0.79&0.32&0.21&0.97&0.93&0.83&0.80\\ 
Loc. so&0.82&0.79&0.31&0.20&0.96&0.93&0.82&0.79\\ 
Loc. mo&-&0.79&0.32&0.21&-&0.93&0.82&0.78\\ 
\hline
\multicolumn{9}{l}{\textit{Drug-protein (ion channels) interaction network}}\\
Global&0.85&0.79&0.31&0.21&0.97&0.93&0.78&0.73\\
Loc. so&0.81&0.80&0.33&0.23&0.97&0.93&0.78&0.74\\ 
Loc. mo&-&0.79&0.33&0.22&-&0.93&0.79&0.74\\ 
\hline
\multicolumn{9}{l}{\textit{Drug-protein (GPCR) interaction network}}\\
Global&0.67&0.53&0.32&0.16&0.95&0.85&0.86&0.81\\
Local so&0.60&0.53&0.33&0.18&0.95&0.84&0.85&0.80\\ 
Local mo&-&0.51&0.31&0.16&-&0.84&0.85&0.81\\ 
\hline
\multicolumn{9}{l}{\textit{Drug-protein (nuclear receptors) interaction network}}\\
Global&0.45&0.29&0.35&0.13&0.84&0.60&0.79&0.66\\ 
Local so&0.43&0.27&0.36&0.12&0.86&0.59&0.80&0.65\\ 
Local mo&-&0.27&0.35&0.12&-&0.59&0.80&0.66\\ 
\hline
\end{tabular} \end{center}

Drug-protein (enzymes) interaction network:
\begin{center}
\includegraphics[width=5cm]{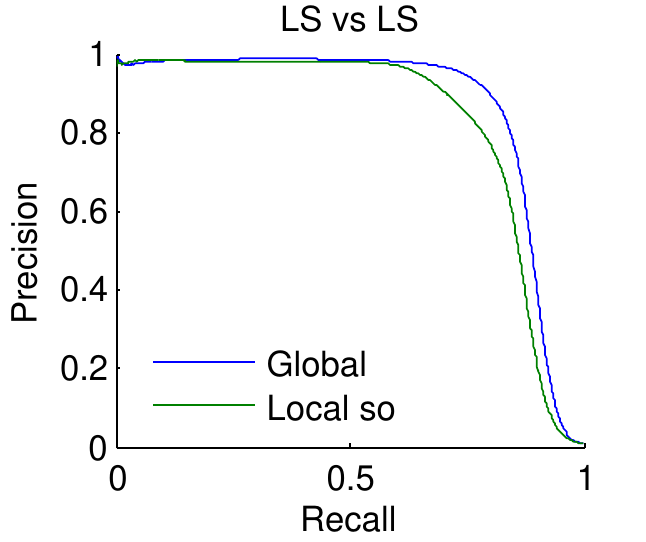}\includegraphics[width=5cm]{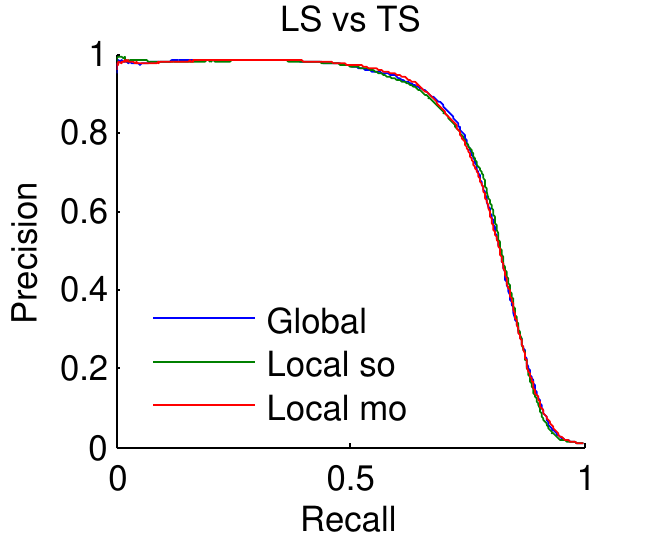} \\
\includegraphics[width=5cm]{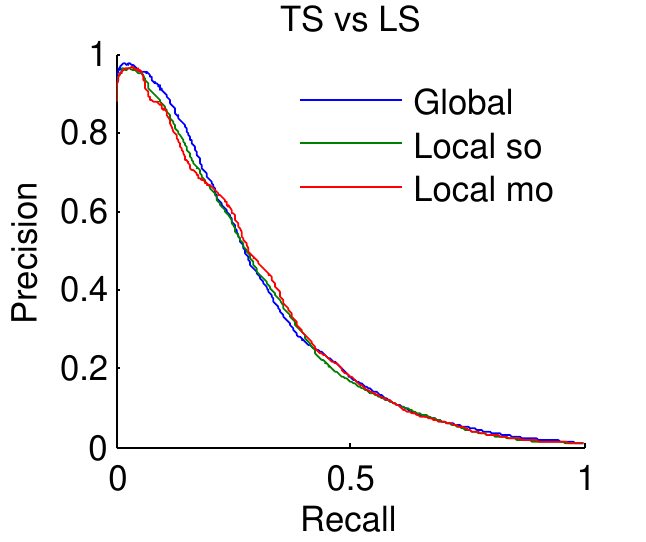}\includegraphics[width=5cm]{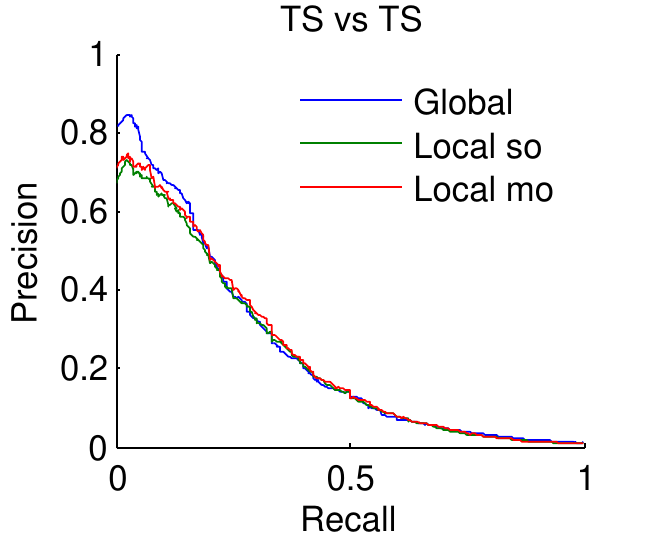}
\end{center}

Drug-protein (ion channels) interaction network:
\begin{center}
\includegraphics[width=5cm]{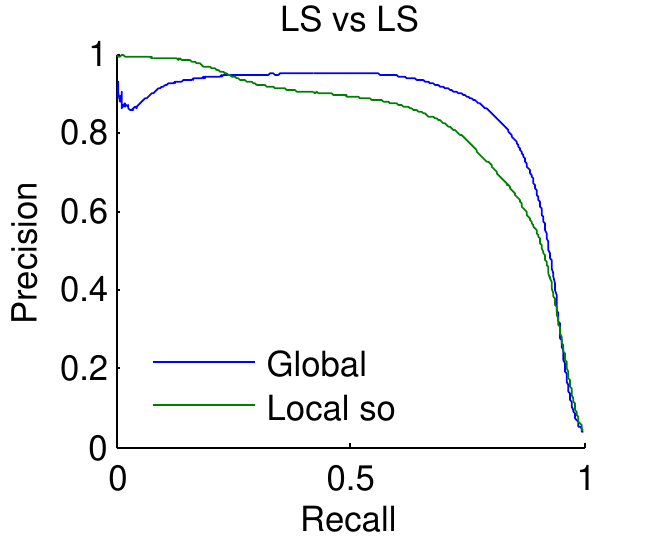}\includegraphics[width=5cm]{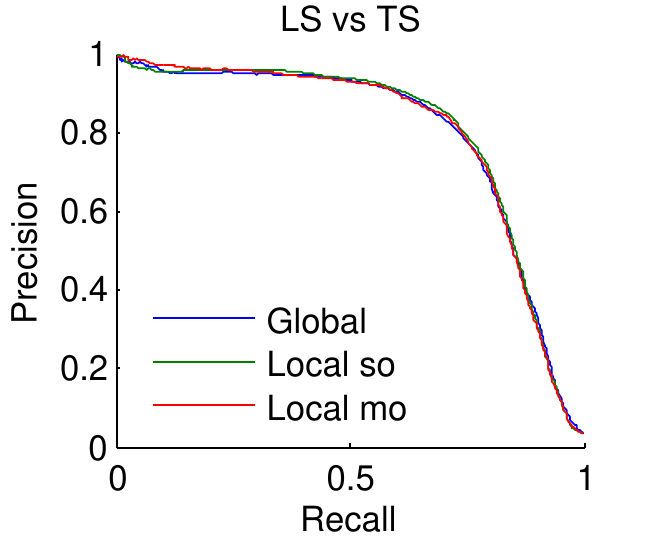} \\
\includegraphics[width=5cm]{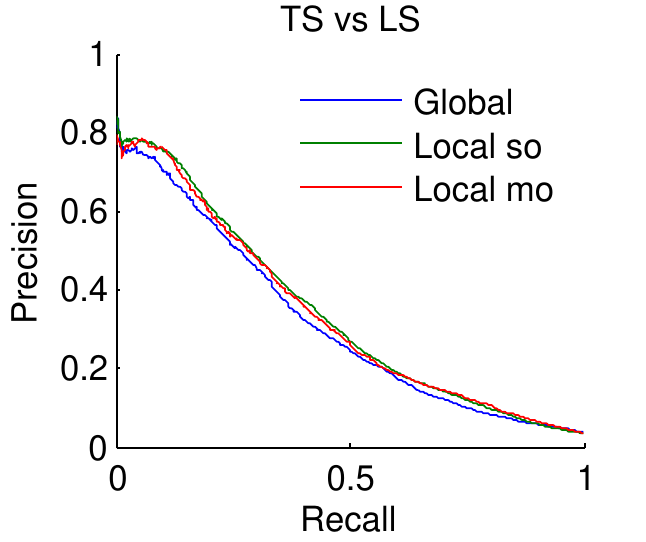}\includegraphics[width=5cm]{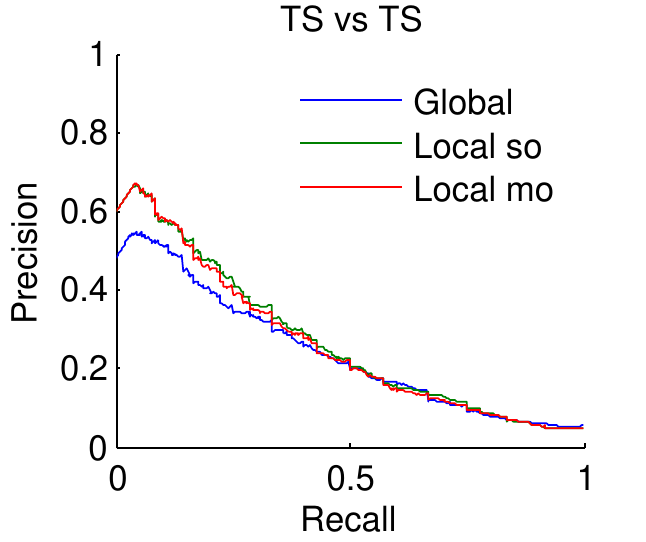}
\end{center}

Drug-protein (GPCR) interaction network:
\begin{center}
\includegraphics[width=5cm]{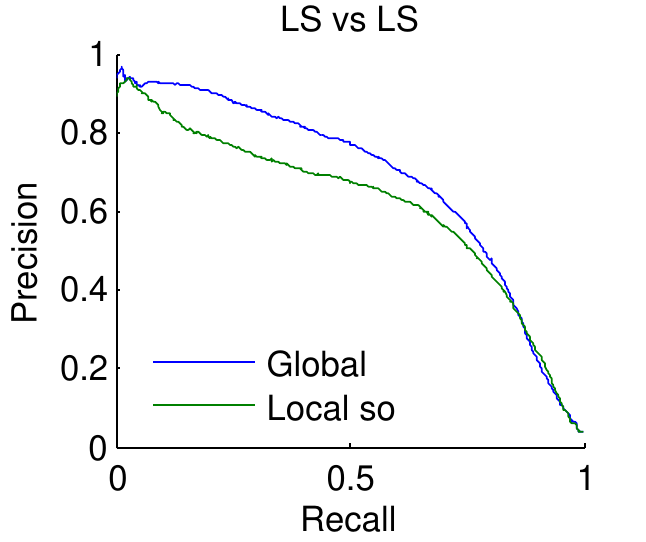}\includegraphics[width=5cm]{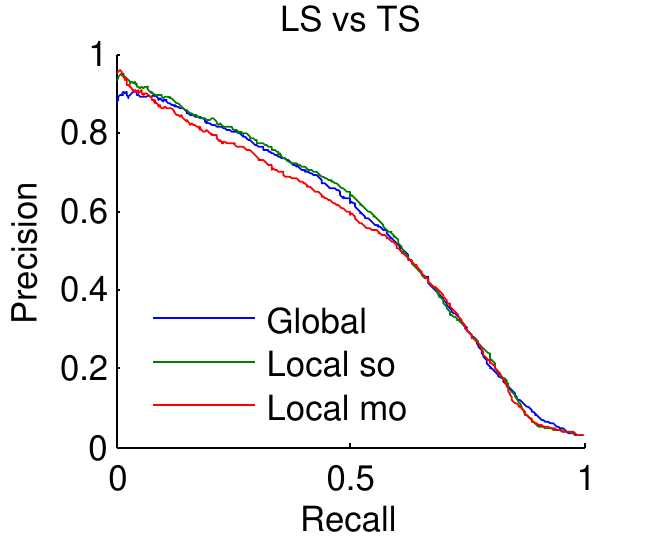} \\
\includegraphics[width=5cm]{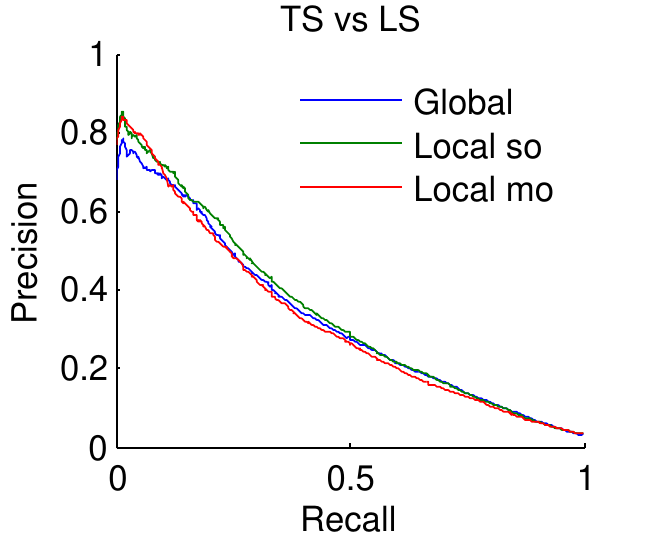}\includegraphics[width=5cm]{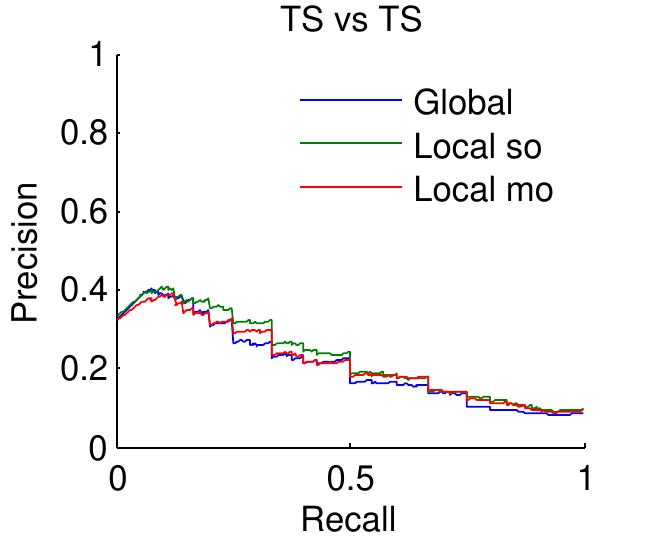}
\end{center}

Drug-protein (nuclear receptors) interaction network:
\begin{center}
\includegraphics[width=5cm]{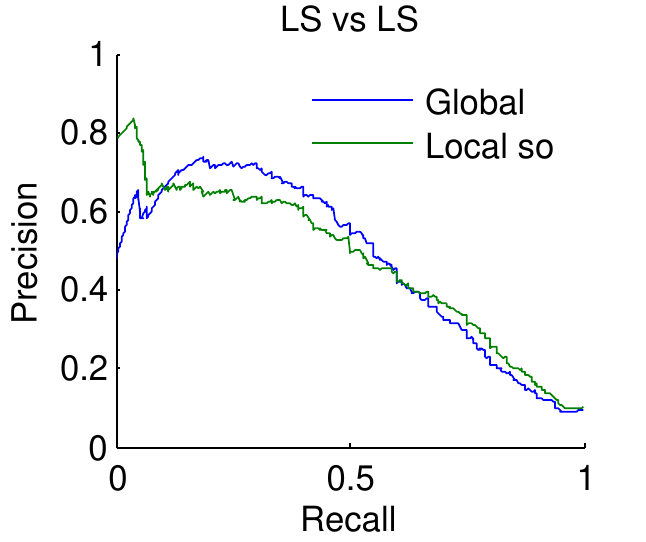}\includegraphics[width=5cm]{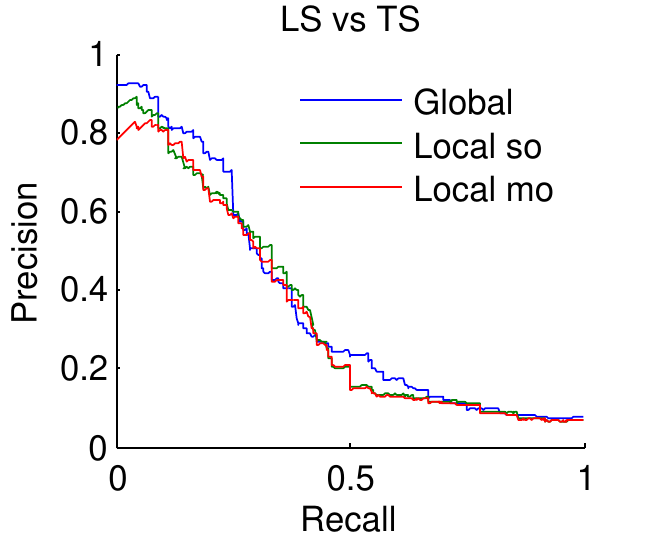} \\
\includegraphics[width=5cm]{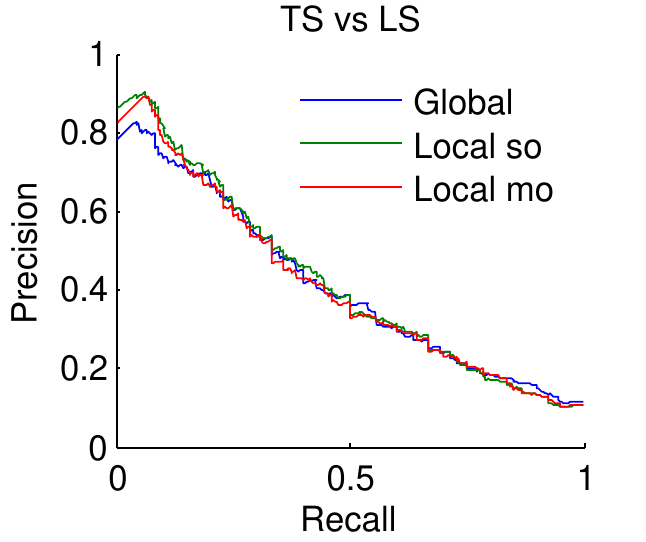}\includegraphics[width=5cm]{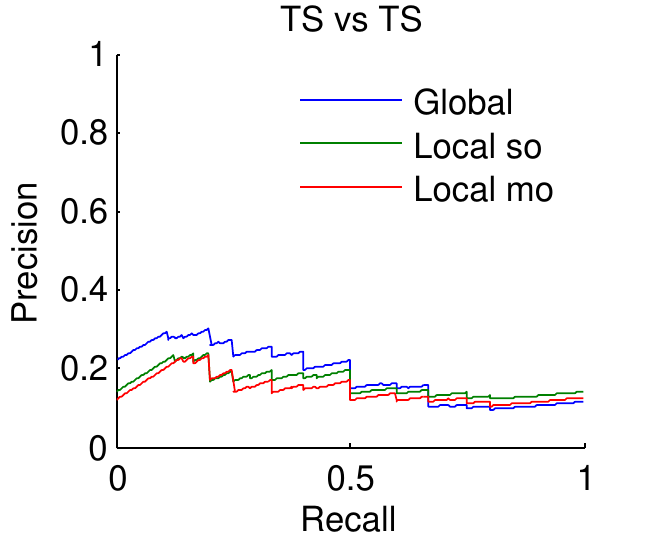}
\end{center}

\paragraph{Comparison with literature.}

\cite{yamanishi2008} and \cite{bleakley2009} use SVM to predict the four classes of drug-protein interaction network. The first one used a kernel regression-based method (KRM): a global approach in which they integrated the chemical and genomic spaces into a unified space. The second one used bipartite local models (BLM) and then did not predict $TS\times TS$ interactions. We compared the AUPR of these two methods with ours, in a 10 times 10-fold CV, in the following Table. Extra-Trees (E-T) is comparable to the other methods, sometimes giving better results (for DPI-I) and sometimes giving less good results (for DPI-N).

\cite{cheng2012} developed three different supervised inference methods, which they tested on the four DPI datasets. The methods are drug-based similarity inference (DBSI), target-based similarity inference (TBSI) and network-based inference (NBI). The last one only use network topology similarity to to infer new targets for known drugs. NBI gives the best performance of the three but has the disadvantage to only be able to predict $LS\times LS$ pairs. Extra-Trees give better or equal results than these three methods, when doing 10 times 10-fold CV. Results are presented in the following Table.


\begin{center}
\begin{tabular}{lllllllll}
&&Method&\multicolumn{3}{l}{Precision-Recall}&&Method&\multicolumn{1}{l}{ROC} \\
&&&LS-LS&LS-TS&TS-LS&&&LS-LS\\ \cline{3-9}
\textit{DPI-E}&& KRM$^1$ &0.83&0.81&0.38&&DBSI$^3$&0.78\\
&&BLM$^2$&0.83&0.81&0.39&&TBSI$^3$&0.90\\
&&&&&&&NBI$^3$&0.97\\
&&E-T&0.87&0.79&0.32&&&0.97\vspace{0.3cm} \\
\textit{DPI-I}&& KRM&0.76&0.81&0.31&&DBSI&0.71\\
&&BLM&0.77&0.80&0.32&&TBSI&0.90\\
&&&&&&&NBI&0.98\\
&&E-T&0.85&0.80&0.34&&&0.97\vspace{0.3cm} \\
\textit{DPI-G}&& KRM&0.67&0.62&0.41&&DBSI&0.76\\
&&BLM&0.65&0.55&0.38&&TBSI&0.75\\
&&&&&&&NBI&0.94\\
&&E-T&0.68&0.55&0.34&&&0.95\vspace{0.3cm} \\
\textit{DPI-N}&& KRM&0.74&0.61&0.51&&DBSI&0.79\\
&&BLM&0.58&0.35&0.40&&TBSI&0.53\\
&&&&&&&NBI&0.84\\
&&E-T&0.48&0.36&0.42&&&0.86\\ \cline{3-9}
\end{tabular} \\
\vspace{0.2cm}
\begin{tabular}{llllll}
$^1$ \cite{yamanishi2008}&&&&&\\
$^2$ \cite{bleakley2009} \\
$^3$ \cite{cheng2012}
\end{tabular}
\end{center}

\bibliographystyle{plain}      
\bibliography{bibli}   


\end{document}